\documentclass{sig-alternate}
\usepackage{amsmath,amssymb}
\usepackage{subfigure}
\usepackage{tikz}
\usepackage{amssymb}
\usepackage{algorithm}
\usepackage[noend]{algpseudocode}
\usepackage{url}
\usepackage{scrextend}
\usepackage{topcapt}
\usepackage{booktabs}
\usepackage{program2}
\usepackage{etex}

\newtheorem{problem}{Problem}

\newtheorem{lemma}{Lemma}

\usepackage{wrapfig}

\newcommand{\wT}{\ensuremath{\widehat{T}}}
\newcommand{\calA}{\ensuremath{\mathcal{A}}}
\newcommand{\error}{{\ensuremath{\textsc{RelError}}}}

\newcommand{\Real}{\ensuremath{\mathbb{R}}}

\newcommand{\minrev}{\ensuremath{\textsc{ActiveCompletion}}}

\newcommand{\movielens}{{\it MovieLens}}
\newcommand{\jester}{{\it Jester}}
\newcommand{\traffic}{{\it Traffic}}
\newcommand{\netflix}{{\it Netflix}}
\newcommand{\boat}{{\it Boat}}
\newcommand{\dataset}[1]{\it {#1}}
 \DeclareMathOperator*{\argmin}{arg\,min}

\newcommand{\targetedicmc}{{\tt Order\&Extend}}

\newcommand{\oecomplete}{{\tt ICMC-O\&E}}

\newcommand{\seqS}{{\tt Sequential}}

\newcommand{\randlmafit}{{\tt LmaFit-Random}}
\newcommand{\randoptspace}{{\tt OptSpace-Random}}
\newcommand{\lmafit}{{\tt LmaFit}}
\newcommand{\optspace}{{\tt OptSpace}}
\newcommand{\icmc}{{\tt ICMC}}

\newcommand{\uc}{{\ensuremath{\mathbb{U}}}}

\newcommand{\spara}[1]{\smallskip\noindent{\bf{#1}}}
\newcommand{\mpara}[1]{\medskip\noindent{\bf{#1}}}

\newcommand{\squishlist}{\begin{list}{$\bullet$}
  { \setlength{\itemsep}{0pt}
     \setlength{\parsep}{3pt}
     \setlength{\topsep}{3pt}
     \setlength{\partopsep}{0pt}
     \setlength{\leftmargin}{1.5em}
     \setlength{\labelwidth}{1em}
     \setlength{\labelsep}{0.5em} } }
\newcommand{\squishend}{
  \end{list}  }

\newcommand{\etal}{{et al.}}

\newfont{\mycrnotice}{ptmr8t at 7pt}
\newfont{\myconfname}{ptmri8t at 7pt}
\let\crnotice\mycrnotice%
\let\confname\myconfname%

\permission{Permission to make digital or hard copies of all or part of this work for personal or classroom use is granted without fee provided that copies are not made or distributed for profit or commercial advantage and that copies bear this notice and the full citation on the first page. Copyrights for components of this work owned by others than ACM must be honored. Abstracting with credit is permitted. To copy otherwise, or republish, to post on servers or to redistribute to lists, requires prior specific permission and/or a fee. Request permissions from Permissions@acm.org.}
\conferenceinfo{KDD'15,}{August 10-13, 2015, Sydney, NSW, Australia.} 
\copyrightetc{\copyright~2015 ACM. ISBN \the\acmcopyr}
\crdata{978-1-4503-3664-2/15/08\ ...\$15.00.\\
DOI: http://dx.doi.org/10.1145/2783258.2783259}

\begin{document}

\title{Matrix Completion with Queries}
%
%
%
%
%

\numberofauthors{3} 
%
\author{
%
%
\alignauthor
Natali Ruchansky\\
       \affaddr{Boston University}\\
       \email{natalir@cs.bu.edu}
\alignauthor
Mark Crovella\\
       \affaddr{Boston University}\\
       \email{crovella@cs.bu.edu}
\alignauthor Evimaria Terzi\\
       \affaddr{Boston University}\\
       \email{evimaria@cs.bu.edu}
}

\date{30 July 1999}

\maketitle

\begin{abstract}

In many applications, e.g., recommender systems and traffic monitoring, 
the data comes in the form of a matrix that is only partially observed and low rank. 
A fundamental data-analysis task for these datasets is
\emph{matrix completion}, where the goal is to accurately infer the
entries missing from 
the matrix.
Even when the data satisfies the low-rank assumption,  classical matrix-completion methods may output completions with significant error -- in that the reconstructed matrix differs significantly from the true underlying matrix. 
Often, this is due to the fact that the information contained in the observed entries is insufficient. In this work, we address this problem by proposing an active version
of matrix completion, where 
queries can be made to the true underlying matrix.
Subsequently, we design {\targetedicmc}, which is the first algorithm to unify a matrix-completion approach and a querying strategy into a single algorithm.  {\targetedicmc} is able identify and alleviate insufficient information 
by judiciously querying a small number of additional entries.
In an extensive experimental evaluation on real-world datasets, 
we demonstrate that
our algorithm is efficient and is able to accurately reconstruct the 
true matrix 
while asking only a small number of queries.

\end{abstract} 

\category{H.2.8 }{Database Management}{Database Applications}
[data mining]
\keywords{matrix completion; recommender systems; active querying}

\section{Introduction}

In many applications the data comes in the form of a low-rank matrix.
Examples of such applications include recommender systems (where entries of the matrix indicated user preferences over items), network traffic analysis
(where the matrix contains the volumes of traffic among
source-destination pairs), and computer vision (image matrices). In many cases,
only a small percentage of the matrix entries are
observed.  For example, the data used in the Netflix prize competition
was a matrix of 
480K users $\times$ 18K movies, but only 1\% of the entries were known.

A common approach for recovering such missing data is called \emph{matrix completion}.
The goal of matrix-completion methods is to accurately infer the values of missing entries, subject to certain
assumptions about the complete
matrix~\cite{candes12exact,candes10power,chen14coherent,jain13lowrank,keshavan10matrix-a,keshavan10matrix-b,wen12solving}.
For a true matrix $T$ with observed values only in a set of positions $\Omega$, 
matrix-completion methods
exploit the information in the observed entries in $T$ (denoted
$T_\Omega$) in order to produce a ``good'' estimate $\wT$ of $T$.
In practice, the estimate
 may differ significantly from the true matrix. 
In particular, this can happen when the observed entries $T_\Omega$ are not adequate to provide sufficient information to produce a good estimate.

In many cases, it is possible to address the insufficiency of $T_\Omega$
by actively obtaining additional observations.  For example, in
recommender systems, users may be asked to rate certain items; in
traffic analysis, additional monitoring points may be installed.  These
additional observations can lead to an augmented $\Omega'$ such that
$T_{\Omega'}$ carries more information about $T$ and can lead to more
accurate estimates $\wT$.  In this \emph{active} setting, the data
analyst can become an \emph{active} participant in data collection by
posing \emph{queries} to $T$.  Of
course such active involvement will only be acceptable if the number of
queries is small.

In this paper, we present a method for
generating a small number of queries so as to ensure that the
combination of observed and queried values provides sufficient
information for an accurate completion; i.e., the $\wT$ estimated using
the entries $T_{\Omega'}$ is significantly better than the one estimated
using $T_\Omega$.  We call the problem of generating a small number of
queries that guarantee small reconstruction error the {\minrev} problem.

The difference between the classical matrix-completion problem and our
problem is that in the former, the set of observed entries is
\emph{fixed} and the algorithm needs to find the best completion given
these entries. In {\minrev}, we are asked to
design both a \emph{completion} and a \emph{querying}
strategy in order to minimize the reconstruction error.  On the one
hand, this task is more complex than standard matrix completion -- since
we have the additional job of designing a good querying strategy. On the
other hand, having the flexibility to ask some additional entries of $T$
to be revealed should yield lower reconstruction error.

At a high level, {\minrev} is related to other recently proposed
methods for active matrix completion,
e.g.,~\cite{chakraborty13active}.  However, existing approaches
identify entries to be queried \emph{independently} of
the method of completion.  In contrast, a strength of our algorithm is that it
addresses completion and querying in an integrated fashion.  

The main contribution of our work is {\targetedicmc}, an algorithm that
simultaneously minimizes the number of queries to ask and produces an
estimate matrix $\wT$ that is very close to the true matrix $T$. The
design of \targetedicmc\ is inspired from recent matrix-completion
methods that view the completion process as solving a sequence of (not
necessarily linear)
systems~\cite{kiraly13algebraic,kiraly12combinatorial,meka09matrix,singer10uniqueness}.
We adopt this general view, focusing on a formulation that
involves only linear systems. Although existing work uses this insight
for simple matrix completion, we go one step further and observe that
there is a relationship between the ordering in which systems are
solved, and the number of additional queries that need to be posed.
Therefore, the first step of {\targetedicmc} focuses on finding a good
ordering of the set of linear systems.  Interestingly, this ordering
step relies on the combinatorial properties of the mask graph, a graph
that is associated with the positions (but not the values) of observed
entries $\Omega$.  

In the second step, {\targetedicmc} considers the
linear systems in the chosen order, and asks queries every time it
encounters a {\em problematic} linear system $Ax=b$.  A linear system
can problematic in two ways: $(a)$ when there
are not enough equations for the number of unknowns, so that the system
does not have a unique solution; $(b)$ when
solving the system $Ax=b$ is numerically unstable {\em given the specific $b$
involved.\/} Note that, as we explain in the paper, this is not the same as simply saying that $A$ is
ill-conditioned; 
part of our
contribution is the design of fast methods for detecting and
ameliorating such systems.

Our extensive
experiments with datasets from a variety of application domains demonstrate that {\targetedicmc} 
requires significantly fewer queries than any other baseline querying
strategy, and compares very favorably to approaches based on well-known
matrix completion algorithms.
In fact, our experiments indicate that {\targetedicmc} is ``almost optimal"
as it gives solutions where the number of entries it queries is generally very close to
the information-theoretic lower bound for completion.

%
%

\section{Related Work}\label{sec:related}

To the best of our knowledge we are the first to pose the problem of constructing an algorithm equipped simultaneously with a completion and a querying strategy.  However, matrix completion is a long studied problem, and in this section we describe the existing work in this area.

\spara{Statistical matrix completion:}
The first methods for matrix completion to be developed
were statistical in nature
\cite{jain2013universal,candes12exact,candes10power,chen14coherent,jain13lowrank,keshavan10matrix-a,keshavan10matrix-b,negahban12restricted,wen12solving}.
Statistical approaches are typically interested in finding a low-rank completion of the partially observed matrix.  
These methods assume a random model for the positions of known entries, and formulate the task as an
optimization problem.   
A key characteristic of statistical methods is that they estimate a completion regardless of whether the information contained in the visible entries is sufficient for completion.
In other words, on any input they output their best estimate, which can have high error.  Moreover, statistical
 methods 
are not equipped with a querying strategy, nor a mechanism to signal when the information is insufficient.

\spara{Random sampling:}
Cand\'{e}s and Recht introduced a threshold on the number of entries needed for accurate matrix completion   \cite{candes12exact}.   Under the assumption of randomly sampled locations of known entries, they prove that an $n_1\times n_2$ matrix of rank $r$ should have at least $m>Cn^{\frac{6}{5}}r\log(n)$ for their algorithm to succeed with high probability, where $n=\max(n_1,n_2)$.  Different authors in the  matrix-completion literature develop slightly different  thresholds, but all are essentially $O(nr\log(n))$ \cite{keshavan10matrix-b,recht2011simpler}.
We point out that achieving this bound in the real world often requires a significantly large number of samples. For example, adopting the rank $r\approx 40$ of top solutions to the Netflix Challenge, over $151$ million entries would need to be queried.

\spara{Structural matrix completion:}
Recently, a class of matrix completion has been proposed, which we call \emph{structural}.
Rather than taking an optimization approach, the methods of structural completion explicitly analyze the information content of the visible entries and
are capable of stating definitively that the observed entries are information-theoretically sufficient for reconstruction \cite{kiraly13algebraic,kiraly12combinatorial,meka09matrix,singer10uniqueness}.  

Structural methods are implicitly concerned with the number of possible completions that are
consistent with the partially observed matrix; this could be infinite, a finite, one, or none.
A key observation shared by all structural approaches is that the number of possible completions
does not depend on the values of the observed entries, but
rather only on their positions. This statement, proved by Kiraly \etal\ \cite{kiraly13algebraic},
means that in our search for good ordering of linear systems we can work solely with the locations of known entries.

The common characteristic between our method and structural methods is that they
also view matrix completion as a task of solving a sequence of (not necessarily linear) systems
of equations where the result of one is used as input to another. In fact, 
Meka {\etal}~\cite{meka09matrix} adopt the same view as ours. However, the key difference
between these works and ours is that we are concerned particularly with the \emph{active} version of the
problem and we need to effectively design both a reconstruction and a querying strategy
simultaneously.



\spara{The active problem:} 
Although active learning has been studied for some time,  work in the active matrix completion area has only appeared recently \cite{chakraborty13active,sutherland13active}.
In both these works,
 the authors are interested in determining which entries need to be revealed in order to reduce error in matrix reconstruction.  Their methods choose to reveal entries with the largest predicted uncertainty based on various measures. Algorithmically, the difference with our work is that the previous approaches construct a querying strategy \emph{independently} of the completion algorithm. 
In fact, they use off-the-shelf matrix completion algorithms for the reconstruction phase, while the strength in our algorithm is precisely its integrated nature of querying and completing.
These methods appear to have other drawbacks.  In the experiments, Chakraborty {\etal} start with partial matrices where 50-60\% of entries are already known -- far greater than that required by our method.  Further, their proposed query strategy does not lead to a significant improvement over pure random querying. While Sutherland {\etal} report low reconstruction error, the main experiments are run over $10\times 10$ matrices, providing no evidence that the methods scale.

\section{Problem Definition}\label{sec:problem}

%
In this section, we describe our setting and provide the problem definition.

\subsection{Notation and setting}

Throughout the paper, we assume the existence of a true matrix $T$ of size $n_1\times
n_2$; $T$ may represent the preferences of $n_1$
users over $n_2$ objects, or the measurements obtained in $n_1$ sites
over $n_2$ attributes.   We assume that the entries of $T$ are real
numbers ($T_{ij} \in \Real $) and that only a subset of these entries 
$\Omega\subset \left\{(i,j)\, \origbar, 1 \leq i \leq n_1, 1\leq j \leq n_2 \right\}$  are
observed.  We refer to the set of positions of known entries
$\Omega$ as the
\emph{mask} of $T$.  When we are referring to the values of the visible
entries in $T$ we will denote that set as $T_\Omega$. 

\begin{figure}
\centering
\includegraphics[scale=.5]{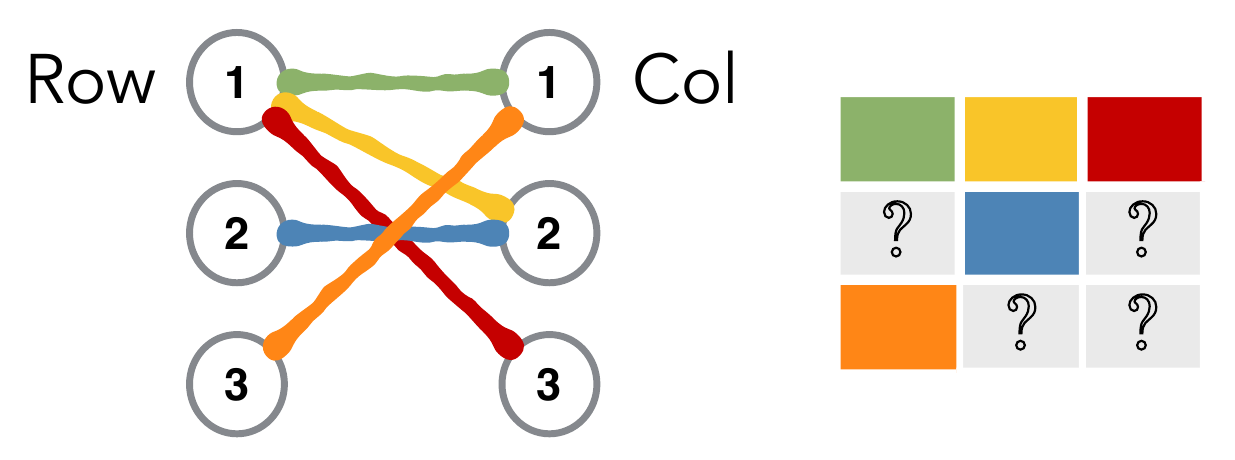} 
\caption{\label{fig:example} The mask graph $G_\Omega$ of mask
 $\Omega = \{(1,1)$, $(1,2)$, $(1,3)$, $(2, 2)$, $(3, 1)\}$.}
\end{figure}

The mask $\Omega$ has an associated \emph{mask graph,}
denoted $G_\Omega$.  The mask graph is a bipartite graph
$G_\Omega = (V_1,V_2,E)$, where $V_1$ and $V_2$ correspond to
the set of nodes in the left and right parts of the graph,
with every node $i \in V_1$ representing a row of $T$ and every $j \in
V_2$ representing a column of $T$.  The edges in $G_\Omega$ correspond to the
positions $\Omega$, meaning $(i, j) \in
E \iff (i, j) \in\Omega$. An example is shown in
Figure~\ref{fig:example}.

Throughout the paper, we will use $\wT_\Omega$ to denote the estimate of 
$T$ that was computed using $T_\Omega$ as input.
Of course, this estimate depends not only on $\Omega$, but also on
the algorithm $\calA$ used for completion.  Therefore, we denote
the estimate obtained by a particular algorithm $\calA$ on input $T_\Omega$
as $\calA(T_\Omega)=\wT_{\calA,\Omega}$. When the algorithm is unspecified or
clear from the context, we will omit it from the subscript.

Finally, we define the \emph{reconstruction error}
of the estimate $\wT_\Omega$ of $T$ as:
\begin{equation}\label{eq:error}
\error(\wT_\Omega) = \frac{\|T - \wT_\Omega\|_F}{\|T\|_F}.
\end{equation}
Recall that for an $n_1\times n_2$ matrix $X$, $\|X\|_F$ (or simply $\|X\|$) is the Frobenius norm defined as
$\|X\|_F =\sqrt{\sum_{i=1}^{n_1}\sum_{j=1}^{n_2}X_{ij}^2}$.

\spara{Low effective rank:} For the matrix-completion problem to be
well-defined, some restrictions need to be placed on $T$. Here, the
\emph{only} assumption we make for $T$ is that it has \emph{low
  effective rank} $r$ $(\ll \min(n_1,n_2))$.  Were $T$ exactly rank $r$, it would
have $r$ non-zero singular values; when $T$ is
\emph{effectively} rank $r$, it has full rank but its top $r$
singular values are significantly larger in magnitude than the rest.
In practice, many matrices obtained through empirical measurements
are found to have low effective rank.


Rather than stipulating low rank, it is often simpler to postulate
that the effective rank $r$ is known.  This assumption is used in
obtaining many theoretical results in the matrix-completion
literature~\cite{candes12exact,kiraly13algebraic,kiraly12combinatorial,negahban12restricted}.
Yet in practice, the important assumption is that of low effective
rank.  Even if $r$ is unknown but required as input to an algorithm, one could try a several
values of $r$ and choose the best performing.
For the rest of the paper, we consider $r$ to be known and omit reference to it when it is understood from context.


\spara{Querying entries:} For simplicity of exposition we discuss our
problem and algorithms in the context of unlimited access to all unobserved entries
of $T$.  However, our results still apply and our
algorithm can still work in the presence of constraints on which
entries of $T$ may be queried.

\subsection{The {\large {\minrev}} problem}

\sloppypar
Given the above, we  define our problem as follows:
\begin{problem}[{\minrev}]\label{problem:theproblem}
Given an integer $r>0$ that corresponds to the effective rank
of $T$ and the values of $T$ in certain positions
$\Omega$,
find a set of additional entries $Q$ to query from $T$
such that for $\Omega'=\Omega\cup Q$, $\error(\wT_{\Omega'})$
as well as $\origbar Q\origbar$ are minimized.
\end{problem}
Note that the above problem definition has two minimization 
objectives: (1) the number of queried entries and (2) the reconstruction error.
In practice we can only solve for one and impose a constraint on the other.
For example, we can impose a \emph{query budget $b$} 
on the number of queries to ask and optimize for the
error. 
Alternatively, one can use \emph{error budget $\epsilon$} to control the error of the 
output, and then minimize for the number of queries to ask. In principle our algorithm
can be adjusted to solve any of the two cases. However, since setting a desired $b$ is more intuitive for our active setting, in our experiments we
do this and optimize for the
error. We will focus on this  
version of the problem (with the budget on the queries) for the majority of the discussion.



\spara{The exact case:} A special case of {\minrev} is when $T$ is
exactly rank $r,$ and the maximum allowed error $\epsilon$ is zero. In
this case, the problem asks for the minimum number of queries required to 
reconstruct $\wT$ that is exactly equal
to the true matrix $T$.  This can only be guaranteed by ensuring that
the information observed in $T_{\Omega'}$ is adequate to restrict the solution space
to a single unique completion, which will then necessarily be
identical to $T$.

Intuitively, one expects that the larger the set
of observed entries $\Omega$, the fewer the number of possible
completions of $T_{\Omega}$. In fact,
Kiraly {\etal}~\cite{kiraly13algebraic} make the fundamental
observation that under certain assumptions, the number of
possible completions of a partially observed matrix does not
depend on the values of the visible entries, but \emph{only} on
the positions of these entries.  This result implies
that the uniqueness of matrices that 
agree with $T_{\Omega}$ is a property of the mask $\Omega$ and not of
the actual values $T_{\Omega}$. 


\emph{Critical mask size:}
The number of degrees of freedom
of an $n_1\times n_2$ matrix of rank exactly $r$ is $r(n_1+n_2-r)$, which we
denote $\phi(T,r)$.
Hence, regardless of the nature of $\Omega$, any solution with $\epsilon=0$
must have $\origbar\Omega'\origbar \ge \phi(T,r)$.  We therefore
call $\phi(T,r)$ the \emph{critical mask size} as it
can be considered
as a (rather strict) \emph{lower bound} on the number of entries that need to be in $\Omega'$ to
achieve small reconstruction error. 

\emph{Empty masks:} For the special case of exact rank and $\epsilon=0$,
if the input mask $\Omega$ is empty, i.e., $\Omega=\emptyset$, then
 {\minrev} can be solved optimally as follows: simply query the entries of $r$ rows and
$r$ columns of $T$. This will require $\phi(T,r)$ queries, which will
construct a mask $\Omega'$ that determines
a unique reconstruction of $T$.
Therefore, when the initial mask $\Omega=\emptyset$, the {\minrev} problem can be solved in polynomial time.

\section{Algorithms}\label{sec:algorithms}

In this section we present our algorithm, \targetedicmc, for addressing
the \minrev\ problem.  

The starting point for the design of {\targetedicmc} 
is the low (effective) rank assumption of $T$. 
As it will become clear, this means that the unobserved entries are related to
the observed entries through a set of linear systems.
Thus one approach to
matrix completion is to solve a sequence of linear systems.
Each system in this sequence
uses observed entries in $T$, or entries 
of $T$ reconstructed by previously solved linear systems
to infer more missing entries.

The reconstruction error of such an algorithm
depends on the quality of the solutions to these linear systems.
As we will show below, each query of $T$ can yield a new equation that can be
added to a linear system.  
Hence, if a linear system has fewer equations than unknowns,
a query must be made to add an additional equation to the system.
Likewise, if solving a system is numerically unstable then a query must
be made to add an equation that will stabilize it.  Crucially, the need
for such queries 
depends on the nature of the solutions 
obtained to linear systems \emph{earlier in the order.}
Thus the order in which systems are solved, and the nature
of these systems are inter-related.    A good ordering 
will minimize the number of ``problematic" systems being encountered.
However, problematic systems can appear even in the best-possible order,
meaning that good ordering alone is insufficient for accurate
reconstruction.

%


At a high level, {\targetedicmc}
operates as follows: first, it finds
a promising ordering of the linear systems. 
Then, it proceeds by 
solving the linear systems in this order.
If a linear system that 
requires additional information is encountered, the algorithm either strategically queries $T$  or
moves the system to the end of the ordering.
When all systems have been solved, $\hat{T}$ is computed
and returned.   
The next subsections describe these steps in detail.

\subsection{Completion as a sequence \\of linear systems}\label{sec:sequence}
In this section we explain the particular linear systems
that the completion algorithm solves, the sequence in which it solves them, and how 
the ordering in which systems are solved affects the quality of the completion.

For the purposes of this discussion, we assume that $T$ is of rank
exactly $r$.  
In this case $T$
can be expressed as
the product of two matrices $X$ and $Y$ of sizes
$n_1\times r$ and $r\times n_2$; that is,
$T=XY$.  
Furthermore, we assume that any subset of $r$ rows of $X$, or $r$ columns
of $Y$, is linearly independent. 
(Later we will
describe how \targetedicmc\ addresses the case when these
assumptions do not hold -- i.e., when $T$ is only
\emph{effectively} rank-$r$, or when an $r$-subset is linearly dependent).
To complete $T,$ it suffices to find such factors $X$ and
$Y$.\footnote{Note that $X$ and $Y$ are not uniquely determined;  any
invertible $r\times r$ matrix $W$ yields new factors $XW^{-1}$ and $WY$
which also multiply to yield $T$. }

\spara{The {\seqS} completion algorithm:}
We start by describing an algorithm we call {\seqS},
which estimates the rows of $X$ and columns of
$Y$.  \seqS\ 
takes two inputs: (1) an ordering $\pi$ over the set of all rows of $X$ and
columns of
$Y$, which we call the \emph{reconstruction order}, and (2) 
the partially observed matrix $T_\Omega$.  

\begin{figure}
\begin{center}
\includegraphics[scale=0.3]{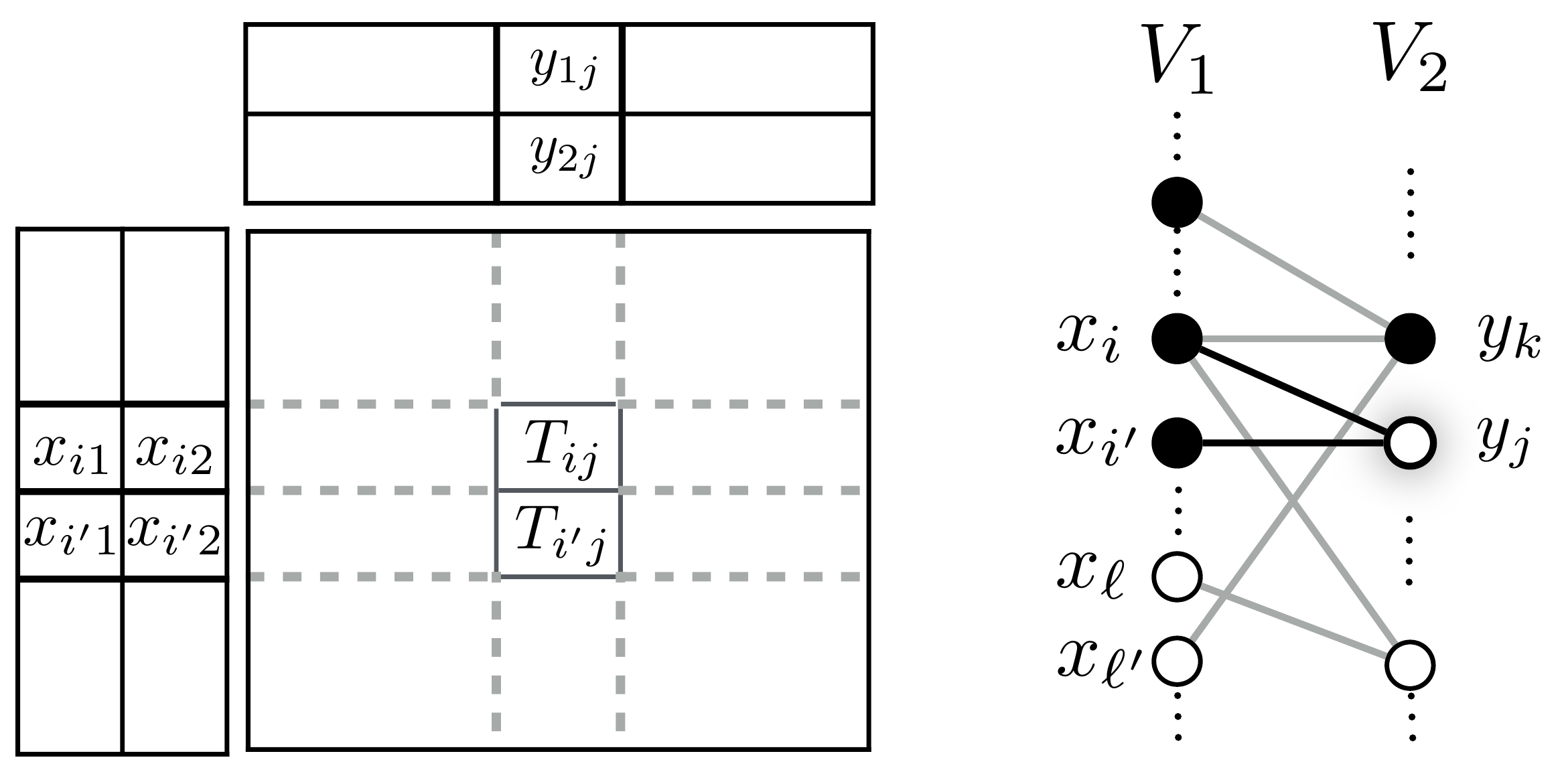}
\caption{\label{fig:icmc-example} An intermediate step of {\seqS} algorithm.}
\end{center}
\end{figure}

To explain how \seqS\ works, consider the example in
Figure~\ref{fig:icmc-example}, where $r = 2$, 
$T$ is on the left and $G_\Omega$ is on the right. 
The factors $X$ and $Y$ are shown on the side of and above
$T$ to convey how their product results in $T$.  The nodes $V_1$ of $G_\Omega$ correspond to the rows of $X$, and nodes $V_2$ to the columns of $Y$.
In this figure, we illustrate an intermediate step of 
{\seqS}, in which the values of the $i$-th and $i'$-th rows of $X$
have already been computed.  
Each entry of $T$ is the inner product of a row of $X$ and a column of
$Y$.  Hence we can represent the depicted
entries in $T$ by the following linear system:
\begin{eqnarray}
T_{ij} & = & x_{i1}y_{1j} + x_{i2}y_{2j}\label{eq:system1}\\
T_{i'j} & =& x_{i'1}y_{1j} + x_{i'2}y_{2j}\label{eq:system2}
\end{eqnarray}
Observe that $x_i$ and $x_{i'}$ are known, and that the edges
$(x_i,y_j)$ and $(x_{i'},y_j)$ corresponding to  $T_{ij}$ and $T_{i'j}$
exist in $G_{\Omega}$.  The only unknowns in (\ref{eq:system1}) and
(\ref{eq:system2}) are $y_{1j}$ and $y_{2j}$, which leaves us with two
equations in two unknowns.  As stated above and by assumption, any
$r$-subset of $X$ or $Y$ is linearly independent; hence one can solve
uniquely for $y_{1j}$ and $y_{2j}$ and fill in column $j$ of $Y$.  

To generalize the example above, the steps of {\seqS} can be partitioned in 
$x$- and $y$-steps; 
at every $y$-step the algorithm 
solves a system of the form 
\begin{equation}\label{eq:linearsystem}
A_x y = t.
\end{equation}
In this system, 
$y$ is a vector of $r$ unknowns corresponding to the values of the column
of $Y$ we are going to compute; $A_x$ is an $r\times r$ fully-known submatrix 
of $X$ and $t$ is a vector of $r$ known entries of $T$ which are located on the same 
column as the column index of $y$.
If $A_x$ and $t$ are known, and $A_x$ is full rank, then $y$ can be computed 
exactly and the algorithm can proceed to the next step.

In the $x$-steps {\seqS} evaluates a row of $X$ using an $r\times r$
already-computed subset of columns of $Y$, and a set of $r$ entries of $T$ from the row of 
$T$ corresponding to the current row of $X$ being solved for.  Following the same notational conventions
as above, the corresponding system becomes $A_yx=t'$.  For simplicity we will 
focus our discussion on $y$-steps; the 
discussion on $x$-steps is symmetric.

\spara{The completion on the mask graph:} 
The execution of {\seqS}  is also captured in the 
mask graph shown in the right part of Figure~\ref{fig:icmc-example}.
In the beginning, no rows or columns have been recovered and all nodes of $G_\Omega$ are white (unknown). As the algorithm
proceeds they become black (known), and this transformation occurs in the order suggested
by the input reconstruction order $\pi$.
Thus a black node 
denotes a row of $X$ or column of 
$Y$ that has been computed. 
In our example, the fact that we can solve for the $j$-th column of $Y$ (using Equations~\eqref{eq:system1} and~\eqref{eq:system2})
is captured by $y_j$'s \emph{two} connections to black/known nodes  (recall $r=2$).
For general rank $r$, 
the $j$-th column of $Y$
can be estimated by a linear system, if in the 
mask graph $y_j$ is connected to at least $r$ already computed  (black) 
nodes.  This is symmetric for the $i$-th row of $X$ and node $x_i$.
Intuitively, this transformation of nodes from black to white is reminiscent of 
an information-propagation process. This analogy was first drawn by Meka {\etal}~\cite{meka09matrix}.

\spara{Incomplete and unstable linear systems:}
As it has already been discussed in the literature~\cite{meka09matrix}, the 
performance of an algorithm like {\seqS} is heavily dependent 
on the input reconstruction order.  Meka {\etal}~\cite{meka09matrix}
have discussed methods for finding a good reconstruction order in the
special case where the 
mask graph has a power-law degree distribution. 
However, even with the best possible reconstruction order {\seqS}
may still encounter linear systems which are either \emph{incomplete} or \emph{unstable}.
Incomplete linear systems are those for which the vector $t$ has some
missing values and therefore the system $A_xy=t$ cannot be solved.
Unstable linear systems are those in which all the entries in $t$ are known, but the 
resulting expression $A_x^{-1}t$ may be
very sensitive to small changes in $t$. These systems raise a numerous problems in the case
where the input $T$ is a noisy version of a rank $r$ matrix, i.e., it is a matrix of 
effective rank $r$. 

In the next two sections we describe how {\targetedicmc} deals with such systems. 
\subsection{Ordering and fixing incomplete systems}\label{sec:ordering}
First, {\targetedicmc} devises an order that minimizes the number of
incomplete systems encountered in the
completion process. 

Let us consider again the execution 
of {\seqS} on the mask graph, and the sequential transformation 
of the nodes in $G_\Omega=(V_1,V_2,E)$ from white to black.  
Recall that in this setting, an incomplete system occurs 
when the node in $G_\Omega$ that corresponds to $y$
is connected to less than $r$ black nodes.

\begin{figure}
\begin{center}
	 \includegraphics[scale=0.35]{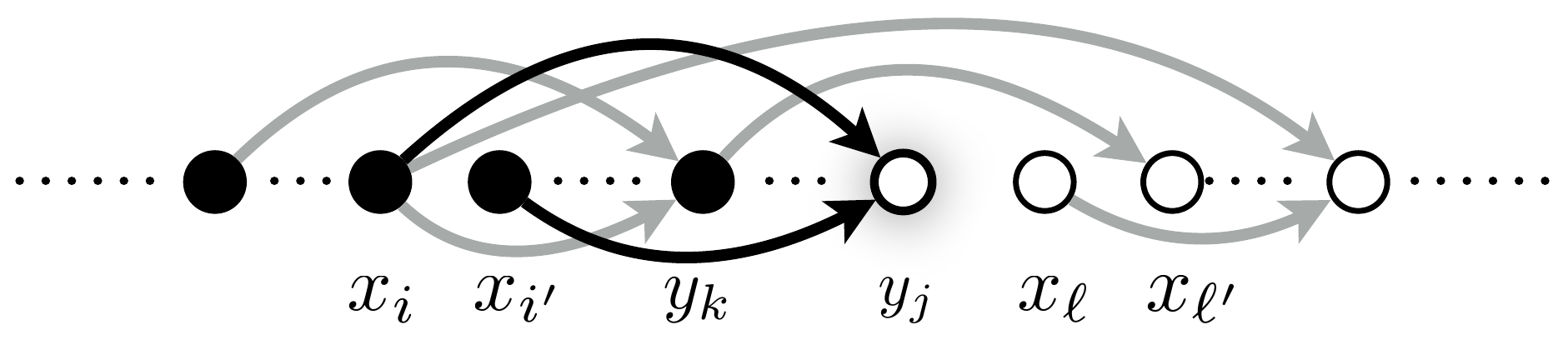}
	 \caption{\label{fig:order} The direction on the edges imposed by the ordering of nodes shown in Figure~\ref{fig:icmc-example}.}
	 \end{center}
\end{figure}

Consider an order $\pi$ of the nodes $V_1\cup V_2$. This order
conceptually imposes a direction on the edges
of $E$; if 
$x$ is before $y$ in that order, then 
$\pi(x)<\pi(y)$, and edge $(x,y)$ becomes directed edge $(x\rightarrow y)$.
Figure~\ref{fig:order} shows this transformation for the mask graph
 in Figure~\ref{fig:icmc-example} and the order implied there.
For fixed $\pi$, a node becomes black if it has at least $r$ incoming edges, i.e., indegree
at least $r$.
In this view, an incomplete system manifests itself by the existence of 
a node that has indegree less
than $r$.
Clearly, if an order $\pi$ guarantees that 
all nodes have $r$ incoming edges, then  there are 
no incomplete systems, and $\pi$ is a \emph{perfect reconstruction order}.

In practice such perfect orders are very hard to find; in most of the cases
they do not exist. The goal of the first step of {\targetedicmc} 
is to find an order $\pi$ that is 
as close as possible to a perfect reconstruction
order. It does so by constructing an order that minimizes the number 
of edges that need to be added so that the indegree of any 
node is $r$.

To achieve this,
the algorithm 
starts by choosing the node from $G_\Omega=
(V_1, V_2,E)$ with the lowest degree. This node is placed last in $\pi$, and
removed from $G_\Omega= (V_1 ,V_2,E)$ along with its incident edges. Of the remaining nodes,
the one with minimum degree is placed in the next-to-last position
in $\pi$, and again removed from $G_\Omega= (V_1, V_2,E)$. This process repeats
until all nodes have been assigned a position in $\pi$.

Next, the algorithm makes an important set of adjustments to $\pi$ by examining each node $u$ in the order it occurs in $\pi$.  
For a particular $u$ the adjustments
can take two forms:
\begin{enumerate}
	\item \emph{if $u$ has degree $\leq r$}: it is repositioned to appear immediately after the neighbor $v$ with the largest $\pi(v)$. 
	\item \emph{if $u$ has degree $>r$}: it is repositioned to appear immediately after the neighbor its neighbor $v$ with the  the
$r$-th smallest $\pi(v)$.
\end{enumerate}

These adjustments aim to construct a $\pi$ such that
when the implied directionality is added to edges, each node has
indegree as close to $r$ as possible.
While it is possible to iterate this adjustment process to further
improve the ordering, in our experiments this showed
little benefit.

Once the order $\pi$ is formed as described above, then the incomplete systems
can be quickly identified: as {\targetedicmc} traverses the nodes of $G_\Omega$ in the order implied by
$\pi$, every time it encounters a node $u$
with in-degree less than 
$r$, it adds edges so that $u$'s indegree becomes $r$;
by definition, the addition of a new edge $(x,u)$ corresponds to querying a missing entry $T_{xu}$ of $T$.

 \subsection{Finding and alleviating unstable systems}\label{sec:unstable}
The incomplete systems are easy to identify -- they correspond to nodes in $G_\Omega$ with degree less than $r$.
However, 
there are other ``problematic" systems which do not appear to be incomplete, yet they are 
 \emph{unstable}. Such systems arise due
to noise in the data matrix or to an
accumulation of error that happens through the sequential system-solving
process.  These systems are harder to detect and alleviate.  We discuss our methodology for this below.

\spara{Understanding unstable linear systems:} Recall that a system $A_xy=t$ is unstable if its solution is very sensitive to the noise
in $t$. 
 To be more specific, consider the system $A_xy=t$, where
 $A_x$ has full rank and $t$ is fully known.
 Recall that the solution
 of this system, $y=A_x^{-1}t$, will be used as part of a subsequent
 system: $A_yx=t'$, where $y$ will become a row
 of matrix $A_y$.
 Let $A_x=U\Sigma V^T$ be the singular value decomposition of $A_x$
 with singular values $\sigma_1\geq \ldots \geq \sigma_r$.
 Now if there is a $\sigma_j$ such that $\sigma_j$ is very small,  then the
 solution to the linear system will be very unstable when the singular
 vector $v_j$ corresponding to $\sigma_j$ has a large projection on $t.$
 This is because in $A_x^{-1}$, the small $\sigma_j$ will be inverted to a very large $1/\sigma_j$.   Thus the inverse operation will cause any component of $t$ that is in the
 direction of $v_j$ to be \emph{disproportionally-strongly expressed}, and any small amount of
 noise in $t$ to be amplified in $y$. 
Thus, unstable systems may be catastrophic
for the reconstruction error of {\seqS} as a single such system
 may initiate a sequence of unstable systems, which
 can amplify the overall error.
 
%

\emph{Unstable vs ill-conditioned systems:}
It is important to contrast the notion of an unstable system 
with that of an ill-conditioned system, which is widely used in the literature.
Recall, that system $A_xy = t$ is ill-conditioned
if  \emph{there exists} a vector $s$ and a small perturbation $s'$ of  $s$, such that
the results of systems $A_xy=s$ and $A_xy'=s'$ are significantly different.
Thus, whether or not a system is ill-conditioned depends only on $A_x$, 
and not on its relationship with any target vector $t$ in particular.
An ill-conditioned system is
also characterized by a large condition number $\kappa(A_x)=\frac{\sigma_1}{\sigma_n}$.
This way of stating ill-conditioning emphasizes that $\kappa(A_x)$ measures a property of $A_x$ and \emph{does not depend}
on $t$.   Consequently (as we will document in Section~\ref{sec:discussion}) the condition
number $\kappa(A_x)$ generates too many false positives to be used for identifying unstable systems.

\spara{Identifying unstable systems:}
To provide a more precise measure of whether a system $A_xy=t$ is unstable, we compute the following quantity:
\begin{equation}\label{eq:localconditionnumber}
\ell(A_x,t)=\| A_x^{-1}\| \frac{\|t\|}{\|y\|}.
\end{equation}
We call this quantity the  \emph{local condition number}, which was
 also discussed by Trefethen and Bau~\cite{trefethen1997numerical}.
 The local condition number is more tailored to our goal as we want to quantify the proneness of a system
 to error with respect to a particular target vector $t$.
 In our experiments, we characterize a system $A_xy=t$ as unstable
if $\ell(A_x,t)\geq \theta$. We call the threshold $\theta$ the 
\emph{stability threshold} and in our experiments we use $\theta =1$. 
Loosely, one can think of this threshold as a way to 
control for the error allowed in the entries of reconstructed matrix. 
Although it is related,
the value of this parameter does not directly translate into
a bound on the {\error} of the overall reconstruction.

\spara{Selecting queries to alleviate unstable systems:}
One could think of dealing with an unstable system 
via regularization, such as ridge regression (Tikhonov Regularization) which was also suggested
 by Meka {\etal}~\cite{meka09matrix}. 
However, for systems $A_xy=t$, such regularization techniques 
aim to dampen the contribution of the singular vector that corresponds to the smallest
singular value, as opposed to boosting the contribution of 
the singular vectors that are in the direction of $t$.  Further, the
procedure can be expressed in terms of only $A_x$ without taking $t$
into account; as we have discussed this is not a good measure for our
approach.

The advantage of our setting is that we can actively query entries from $T$.
Therefore, our way of dealing with this problem is by 
adding a direction to $A_x$ (or as many as are needed until there are $r$ strong ones). 
We do that by extending our system from $A_xy=t$ to 
$\begin{bmatrix} A_x \\ \alpha	\end{bmatrix} \tilde{y}=\begin{bmatrix} t\\ \tau	\end{bmatrix}$.
Of course, in doing so we implicitly shift from looking for an exact
solution to the system, to looking for a least-squares solution.

Clearly $\alpha$ cannot be an arbitrary vector. It must be an already computed row
of $X$, it should be independent of $A_x$, 
and it must boost a direction in $A_x$ which is poorly expressed and also in the direction of $t$.
Given the intuition we developed above, we iterate over all previously computed rows of $X$ that are not in $A_x$, and set each row
as a candidate $\alpha$.
Among all such $\alpha$'s we pick $\alpha^\ast$ as the one with the smallest $\ell(A_x,t)$, and use it to extend  $A_x$ to
$\begin{bmatrix} A_x \\ \alpha^\ast\end{bmatrix}$.

\begin{algorithm}
\begin{algorithmic}
\caption{\label{algo:local}The {\tt local\_condition} routine}
\Statex {{\bf Input:} $C,A_x,\alpha,t$}
\State {$D=C-\frac{C\alpha^T\alpha C}{1+\alpha C\alpha^T}$ }
\State {$\tilde{A_x} = \begin{bmatrix}A_x\\\alpha\end{bmatrix}$}
\State {$\tau = \texttt{Random}\left(T(i,:),T(:,j)\right)$}
\State {$\tilde{t} = \begin{bmatrix}t\\\tau\end{bmatrix}$}
\State {$\tilde{y}=D\tilde{A_x}\tilde{t}$}
\State {{\bf return} $\|D\tilde{A_x}\|{\|\tilde{t}\|\over \|\tilde{y}\|}$}
\end{algorithmic}
\end{algorithm}

\begin{algorithm}
\begin{algorithmic}
\caption{\label{algo:selection}The {\tt Stabilize} routine}
\Statex{{\bf Input:} $A_x,t,\theta$}
\State $j:$ the column of $Y$ being computed
\State  $C = (A_x^T A_x)^{-1}$
 \For {$i \in \{\text{Computed rows of X}\}$}  
 \State {$\alpha_i=X(i,:)$}
\If {$X(i,:)\text{ not in } A_x$} 
\State $c(i)=\texttt{local\_condition}(C, A_x, \alpha_i, t, \tau)$ 
\EndIf
\EndFor
\State {$i^\ast=\argmin_i c(i)$, $\alpha^\ast = X(i^\ast,:)$}
\If {$ c(i^\ast) < \theta$ }
\State {\bf return} $(i^\ast,j),\alpha^\ast$
\EndIf
\State {\bf return} null
\end{algorithmic}
\end{algorithm}

\emph{Querying $T$ judiciously:} Although the above procedure is
conceptually clear, it raises a number of practical 
issues.
If the system $A_xy=t$ solves for the $j$-th column of matrix $Y$,
then every time we try a different $\alpha$, which suppose is the already-computed $X(i,:)$, then the corresponding $\tau$ must be
the entry $T_{ij}$.  Since $T_{ij}$ is not necessarily an observed entry, this would require a query even for rows $\alpha\neq \alpha^\ast$, which is clearly 
a waste of queries since we will only pick one
$\alpha^\ast$. Therefore, instead of querying the unobserved  
values of 
$\tau$, {\targetedicmc} simply uses random values following the distribution of the
values observed in the $i$-th row and $j$-th column of $T$. Once $\alpha^\ast$
is identified, we only query the value of $\tau$ corresponding to row
 $\alpha^\ast$ and column $j$.

If there is no $\alpha^\ast$ that leads to a system with 
local condition number below our threshold, we  postpone solving this
system by moving the corresponding node of the mask graph to the end of the 
order $\pi$.

\spara{Computational speedups:} From the computational point of view, the above approach requires computing a matrix inversion per $\alpha$. With a cubic algorithm for matrix inversion,
this could induce significant computational cost. 
However, we observe that 
this can be done efficiently as 
all the matrix inversions we need to perform are for matrices that differ
only in their last row -- the one occupied by $\alpha$.

Recall that the least-squares solution of the system
$A_xy=t$ is $y  =  (A_x^TA_x)^{-1}A_x^Tt$.
Now in the extended system $\begin{bmatrix} A_x \\ \alpha	\end{bmatrix} \tilde{y}=\begin{bmatrix} t\\ \tau	\end{bmatrix}$ or $\tilde{A_x}\tilde{y}=\tilde{\tau}$, the corresponding solution is
$\tilde{y}  =  (\tilde{A_x}^T\tilde{A_x})^{-1}\tilde{A_x}^T\tilde{t}$.  Observe that we can write:
\[  \tilde{A_x}^T\tilde{A_x} = \begin{bmatrix} A_x^T \  \alpha^T	\end{bmatrix} \begin{bmatrix} A \\ \alpha	\end{bmatrix} = A_x^TA_x+\alpha^T\alpha.\]

Thus, $\tilde{A_x}^T\tilde{A_x}$ can be seen  as a  rank-one update to
$A_x^TA_x$. In such a setting the
Sherman-Morrison Formula~\cite{golub12matrix} provides a way to
efficiently calculate $ D = (\tilde{A_x}^T\tilde{A_x} )^{-1}$ given
$C=(A_x^TA_x)^{-1}$.   The details are shown in Algorithm~\ref{algo:local}.
Using the Sherman-Morrison Formula we can find $\tilde{y}$ 
via matrix multiplication, which requires $O(r^2)$ for at most $n=\max\{n_1,n_2\}$ candidate queries. Since the values of $r$ we encounter in real datasets are small constants
(in the range of 5-40), this running time is small.

The pseudocode of this process is shown in Algorithm~\ref{algo:selection}. The process of selecting the right
entry to query is summarized in the {\tt Stabilize} routine.  Observe that {\tt Stabilize}
either returns the entry to be queried, or if there is no entry that can lead to a stable systems it returns null.  In the latter
case the system is moved to the end of the order.

\subsection{Putting everything together}\label{sec:everything}
Given all the steps we described above we are now ready to summarize
{\targetedicmc} in Algorithm~\ref{algo:targetedicmc}.

\begin{algorithm}
\begin{algorithmic}
\caption{\label{algo:targetedicmc}The {\targetedicmc} algorithm}
\Statex{{\bf Input:} $T_\Omega, r,\theta$}
\State Compute $G_\Omega$
\State  Find ordering $\pi$ (as per Section~\ref{sec:ordering})
\For {$A_xy=t$ (corresponding to the $j$-th column of $Y$) encountered in $\pi$}
\State $\text{solve\_system} = true$
\If{$A_xy=t$ is incomplete}
\State {Query $T$ and complete $A_xy=t$}
\EndIf
\While{$\texttt{local\_condition}(A_x,t)>\theta$}
\If{$\left\{(i^\ast,j),\alpha^\ast\right\}=\texttt{Stabilize}(A_x,t,\theta)\neq \text{null}$}
\State {$A_x = \begin{bmatrix}A_x\\\alpha^\ast\end{bmatrix}$}
\State {$t = \begin{bmatrix}t\\T(i^\ast,j)\end{bmatrix}$}
\Else 
\State {move $A_xy=t$ to the end of $\pi$}
\State $\text{solve\_system}=false$
\State break
\EndIf
\EndWhile
\If {solve\_system}
\State $Y(:,j) = y = A_x^\dagger t$ (using least squares)
\EndIf
\EndFor
\State {\bf return} \wT = XY
\end{algorithmic}
\end{algorithm}

{\targetedicmc} constructs the rows of $X$ and columns of $Y$ in the order prescribed by 
$\pi$ -- the pseudocode shows the construction of columns of $Y$, but it is symmetric for the rows of $X$.
For every linear system the algorithm encounters, it completes the system if it is incomplete and tries to 
make it stable if it is unstable. When a complete and stable version of the system is found, the system is solved using
least squares. Otherwise, it is moved to the end of $\pi$.

\spara{Running time:} The running time of {\targetedicmc} consists of the
 time to obtain the initial
ordering, which using the algorithm of Matula and Beck~\cite{matula1983smallest} is $O\left(n_1+n_2\right)$, plus the time to detect
and alleviate incomplete and unstable systems.  Recall that for each unstable system we compute an inverse $O\left(r^3\right)$ and check $n$ candidates $O\left(r^3+r^2n\right)$.
Thus the overall running time of 
our algorithm is 
 $O\left((n_1+n_2)+N\times(r^3+r^2n)\right)$, where $N$ is the number 
of unstable system the algorithm encounters. In practice, the closer a matrix is to 
being of rank exactly $r$, the smaller the number of error prone systems it encounters and 
therefore the faster its execution time.\footnote{Code and information are available at \url{http://cs-people.bu.edu/natalir/matrixComp}}



\spara{Partial completions:}
If the budget $b$ of allowed queries is not adequate to resolve
the incomplete or the 
unstable systems, then {\targetedicmc} will 
output $\wT$ with only a portion of the entries completed. The entries that remain unrecovered  
are those for which the algorithm claims inability to produce a
good estimate. From the practical viewpoint this is extremely useful information as the 
algorithm is able to inform the data analyst which entries it was not able to reconstruct from the observations in $T_{\Omega}$.

\section{Experiments}\label{sec:experiments}

 \begin{figure*}
	 \begin{center}
	 \subfigure[{\dataset{Traffic1}} ]{
	 \includegraphics[scale=0.3]{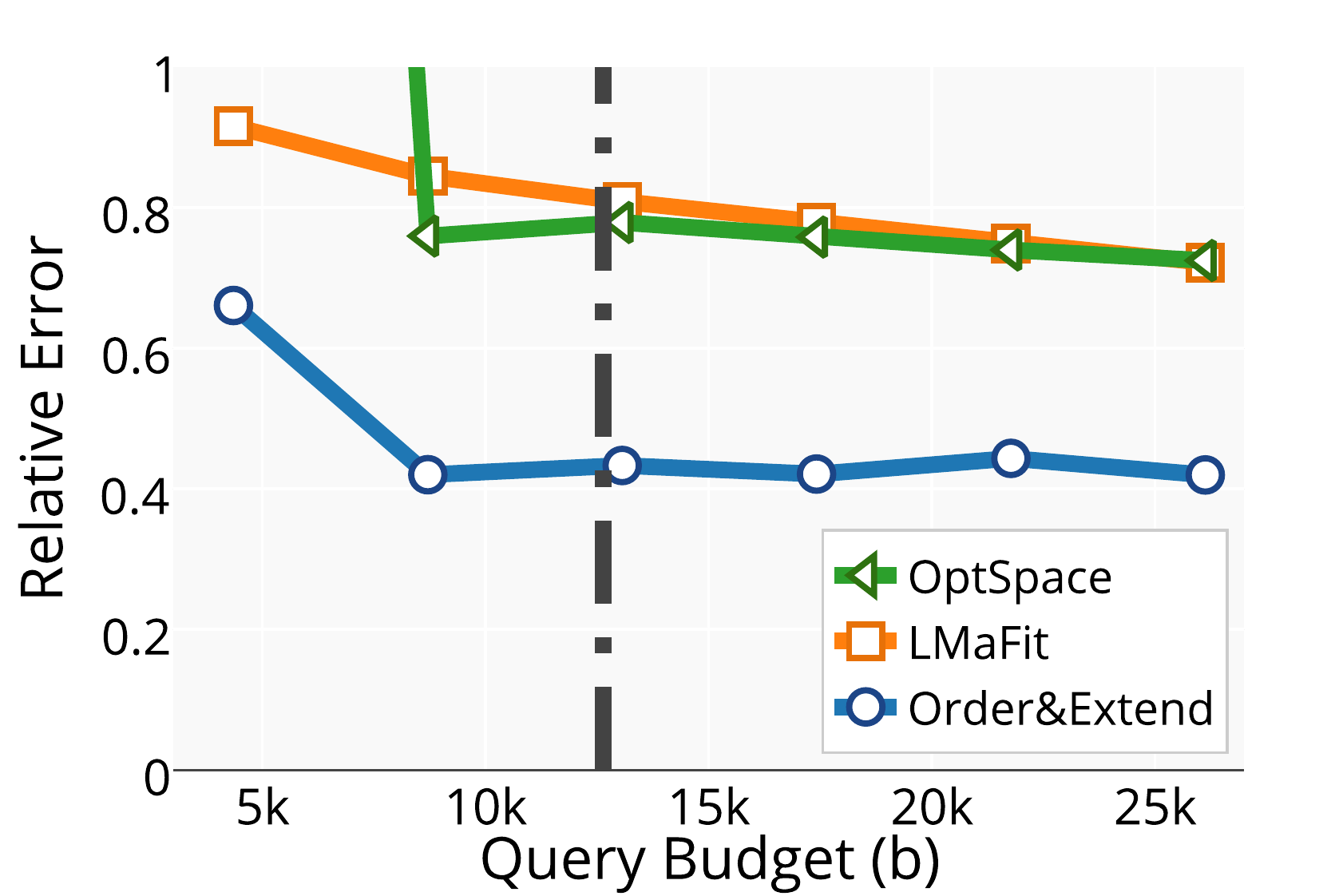}
	 }
	 \subfigure[{\dataset{Traffic2}}  ]{
	 \includegraphics[scale=0.3]{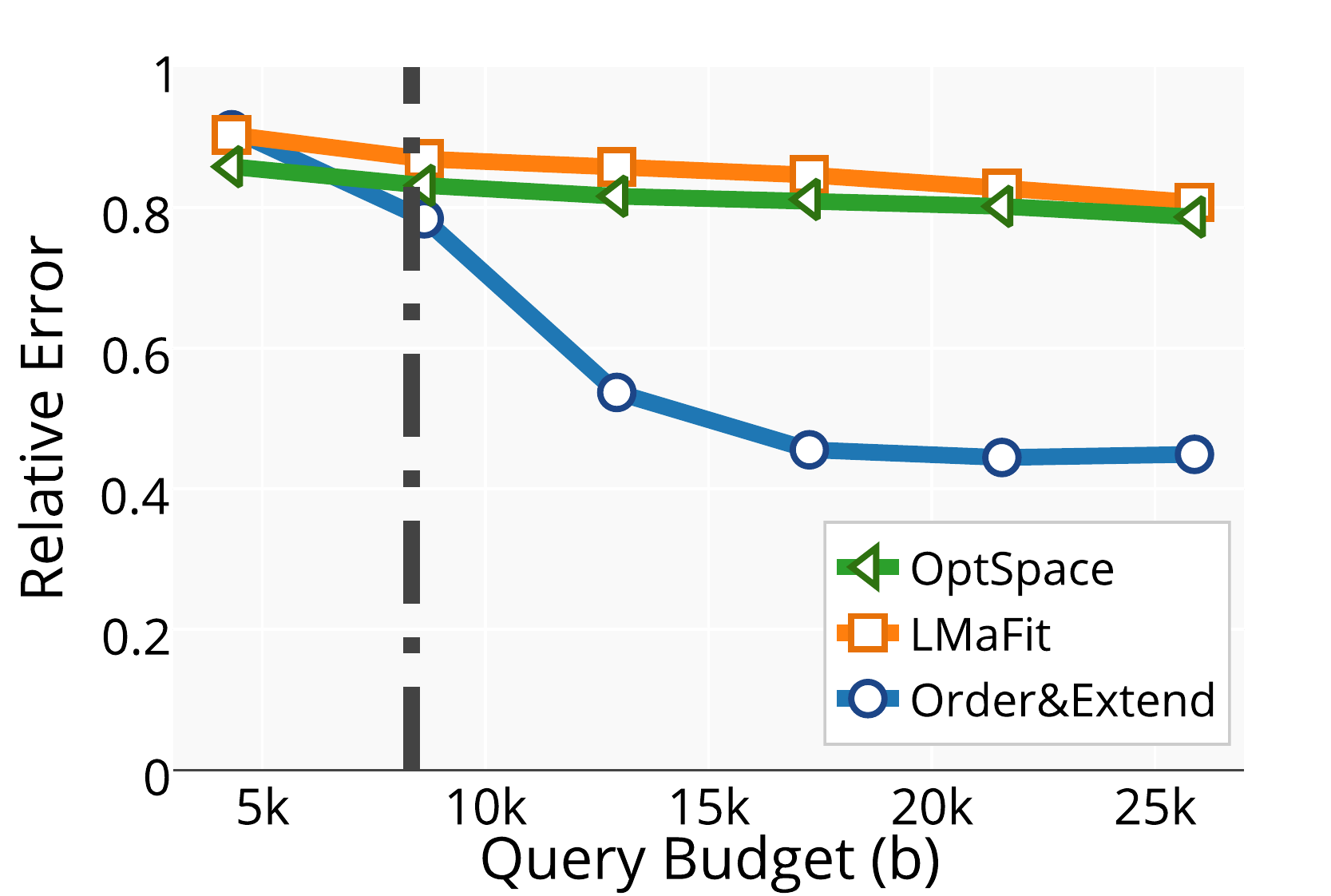}
	 }
	 \subfigure[{\dataset{Boat}}  ]{
	 \includegraphics[scale=0.3]{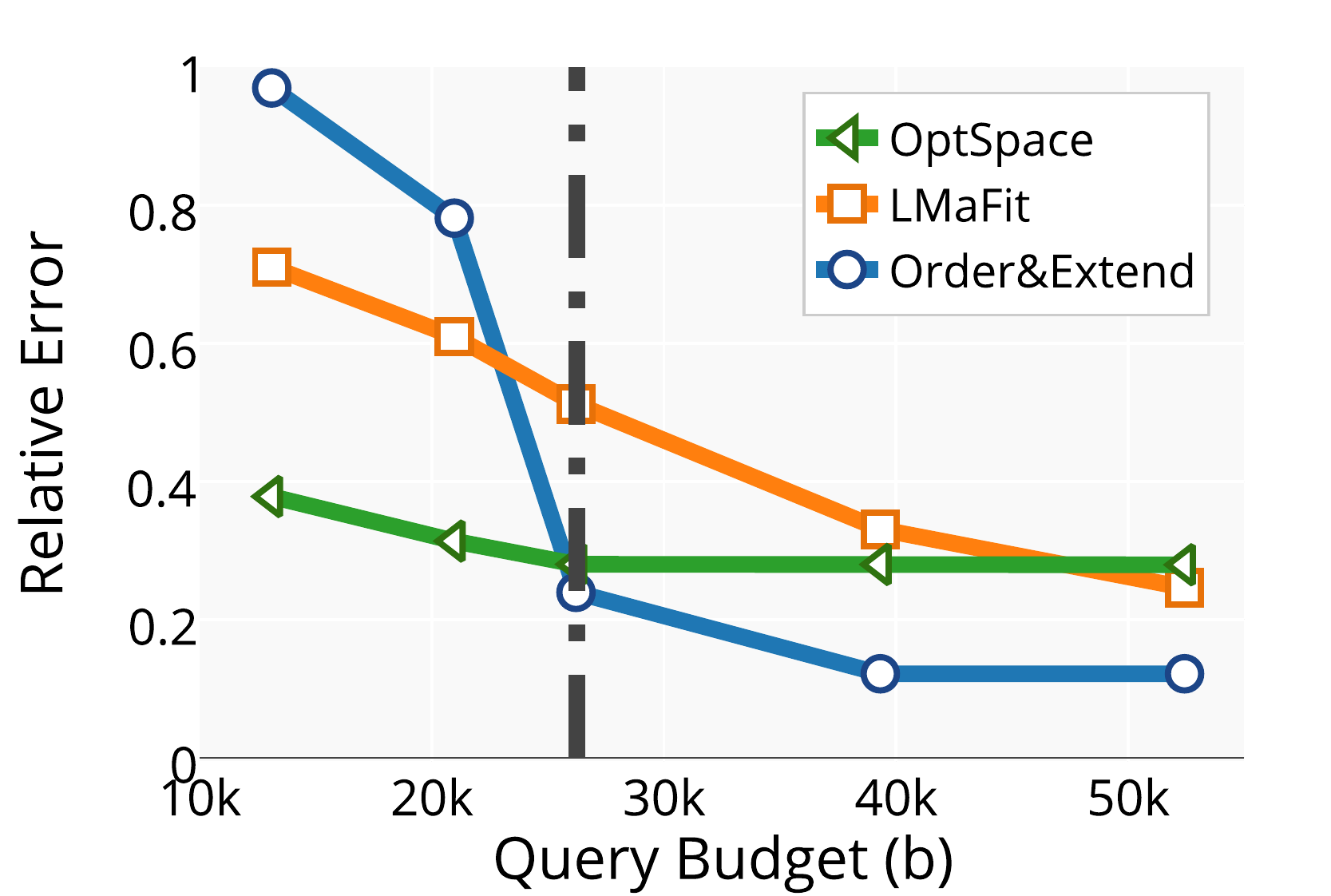}
	 }
	 \subfigure[{\dataset{Latency1}}  ]{
	 \includegraphics[scale=0.3]{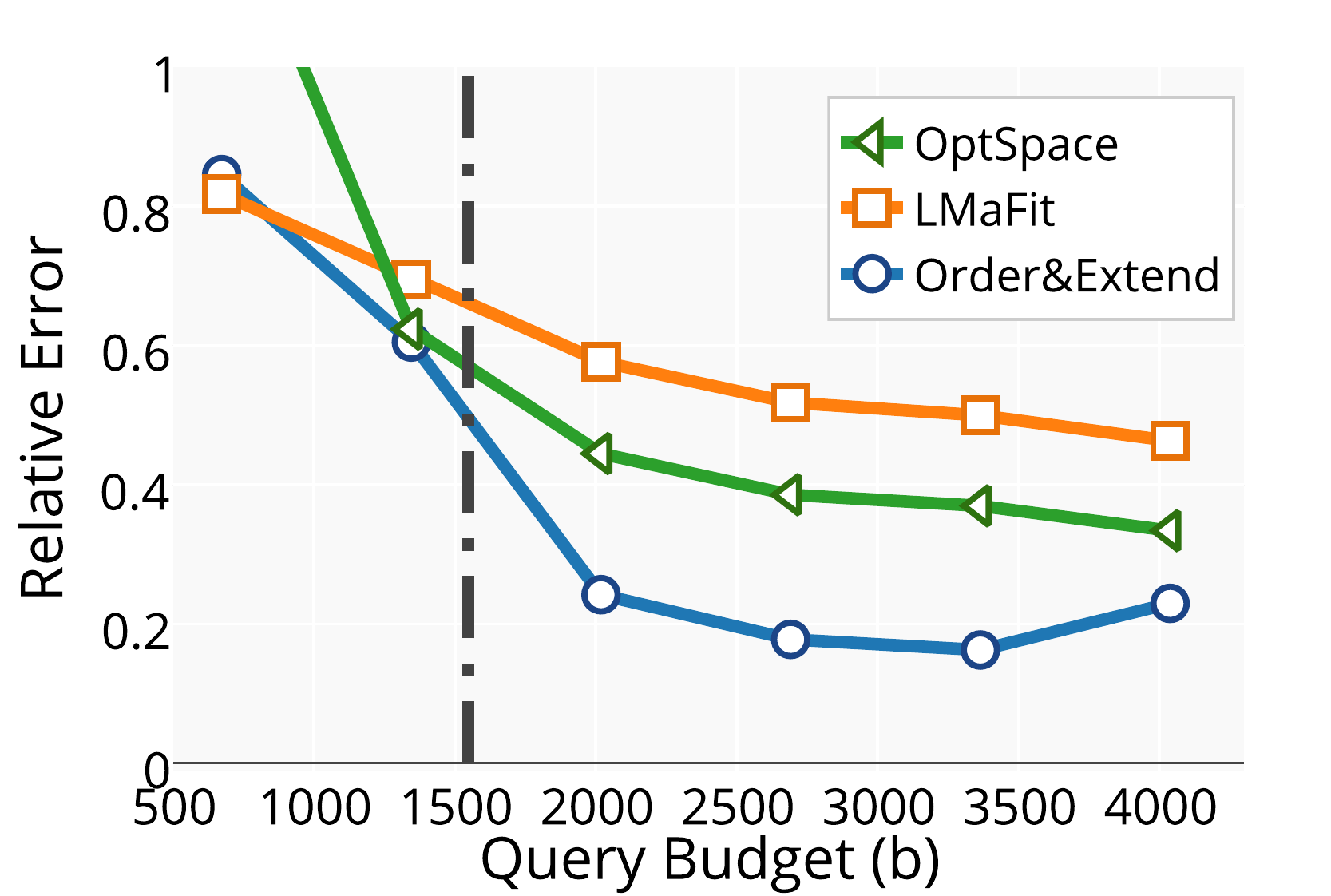}
	 }
	 \subfigure[{\dataset{Latency2}}  ]{
	 \includegraphics[scale=0.3]{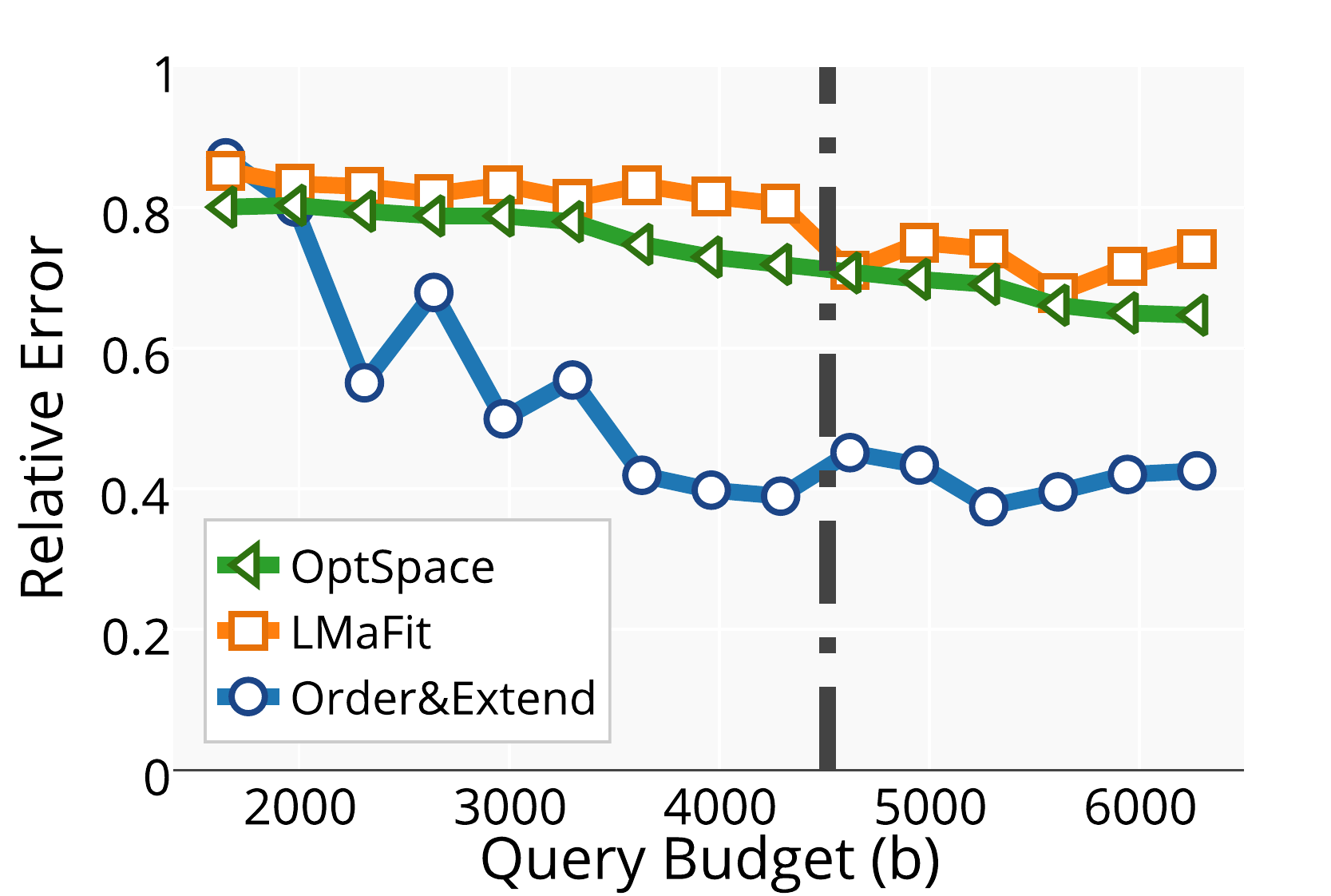}
	 }
	 \subfigure[{\dataset{Jester}}  ]{
	 \includegraphics[scale=0.3]{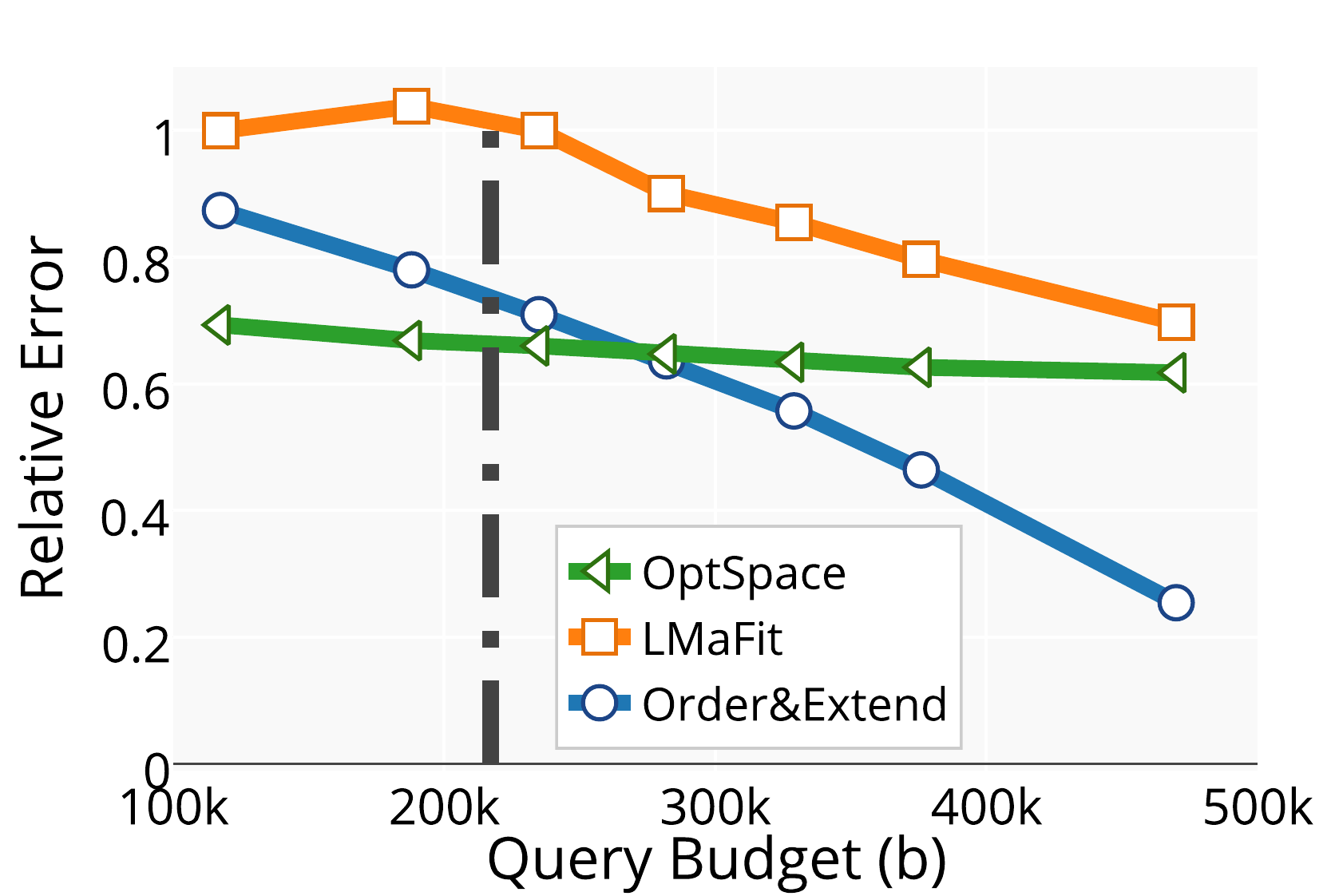}
	 }

	 \end{center}
	\caption{\label{fig:errorReal} {\error} of completion achieved by {\targetedicmc}, {\lmafit} and {\optspace} on datasets with \emph{approximate} rank; $x$-axis: query budget $b$; $y$-axis: {\error} of the completion.}
	 \end{figure*}
	 
In this section we 
experimentally evaluate the performance of 
{\targetedicmc} both in terms of reconstruction error 
as well as the number of queries it makes.
Our experiments show that across all datasets \targetedicmc\ requires very few queries 
to achieve a very low reconstruction error. All other baselines we compare against
require many more queries for the same level of error, or can ever achieve the same
level of reconstruction error.

\spara{Datasets:}
We experiment on the following nine real-world datasets, taken from a variety of applications.

{\movielens}: This dataset contains ratings of users for movies as appearing in the 
 MovieLens website.\footnote{Source~\url{http://www.grouplens.org/node/73}.}
The original dataset has size $6\,040\times 3\,952$ and 
only $5\%$ of its entries are observed. For our experiments we obtain a denser matrix 
of size  $4\,832\times 3\,162$.
 
{\netflix}: This dataset  also contains user movie-ratings, but from the Netflix website. The dataset's original size is $480\,189\times17\,770 $ with $1\%$ of observed entries. Again we focus on a submatrix with higher percentage of observe entries
and size $48\,019\times 8\,885$.

{\jester}: This dataset corresponds to a 
collection of user joke ratings obtained for joke recommendation on the Jester website.\footnote{Source~\url{http://goldberg.berkeley.edu/jester-data/}}  
For our experiments we use the whole dataset 
with size $23\,500\times 100$ with $72\%$ of its entries being observed.

{\boat}: This dataset corresponds to a fully-observed  black and white image of size $512\times 512$.
 
{\traffic}: This is a set of four datasets; each is part of a traffic matrix from a large
 Internet Service Provider where rows and the columns 
 are source and destination prefixes (i.e., groups of IP addresses), and each 
entry is the volume of traffic between the corresponding source-destination pair. 
The largest dataset size  $7\,371\times 7\,430$ 
and $0.1\%$ of its entries are observed; we call this
{\dataset{TrafficSparse}}.  The other two are fully-observed of sizes
$2\, 016\times 107$, and $2\, 016\times 121$; we call these
{\dataset{Traffic1}} and {\dataset{Traffic2}}.\footnote{\label{mark}Source~\url{https://www.cs.bu.edu/~crovella/links.html}}

{\dataset{Latency}}: Here we use two datasets consisting of Internet
network delay measurements.  Rows and columns are hosts, and each entry
indicates the minimum ping delay among a particular time window.  The
datasets are fully-observed and of sizes $116\times 116$, and $869\times
19$; we call these {\dataset Latency1} and {\dataset Latency2}.\footref{mark}

%
%



\spara{Baseline algorithms:}
We compare the performance of our algorithm to two state-of-the-art matrix-completion algorithms, \optspace\ and \lmafit\ . 

{\optspace}: An SVD-based algorithm introduced by Keshavan 
{\etal}\cite{keshavan10matrix-a}.  The algorithm centers around a convex-optimization step that aims to minimize the disagreement of the estimate $\wT$
on the initially observed entries $T_\Omega$. We use the original implementation of
{\optspace}.\footnote{\url{http://web.engr.illinois.edu/~swoh/software/optspace/code.html}}

 {\lmafit}: A popular alternating least-squares method for matrix completion~\cite{wen12solving}.  In our experiments we use the original implementation of this algorithm
provided by Wen et. al.\ \footnote{\url{http://lmafit.blogs.rice.edu/}}, and in particular the version where the rank $r$ is provided, as we observed it to perform best.


As neither {\optspace} nor {\lmafit} are algorithms for active completion,
 we set up our experiment as follows: 
first, we run \targetedicmc\ on $T_{\Omega_0}$, which asks a budget of $b$ queries.  Before feeding $T_{\Omega_0}$ to \lmafit\ and \optspace\ we extend it with $b$ \emph{randomly} chosen queries.  
In this way both algorithms query the same number of additional entries.
A random distribution of observed entries has been proved to be (asymptotically) optimal for
statistical methods like {\optspace} and {\lmafit}~\cite{candes12exact,keshavan10matrix-a,wen12solving}. Therefore, picking randomly distributed $b$ additional 
entries is the best querying strategy for these algorithms, and we have also verified 
that experimentally. 

\subsection{Methodology}
For all our experiments, the ground-truth matrix $T$ is known but not fully revealed to the algorithms. The input to the algorithms consists of
an \emph{initial 
mask} $\Omega_0$, the observed matrix $T_{\Omega_0}$, and a budget $b$ on the number of queries they can ask.
Each algorithm $\calA$ outputs an estimate $\wT_{\calA,\Omega_0}$ of $T$.

\spara{Selecting the input mask $\Omega_0$:}
The initial mask $\Omega_0$, with cardinality $m_0$ is selected 
by picking $m_0$ entries uniformly at random from the ground-truth matrix $T$.\footnote{We also test other sampling distributions, but the results are the same as the ones we report here and thus omitted.}
The cardinality $m_0$ is selected so that $m_0>0$ and $m_0<\phi(T,r)$;
usually we chose $m_0$ to be $\approx 30-50\%$ of $\phi(T,r)$.
The former constraint guarantees that the input is not trivial, while the latter guarantees that additional queries are definitely needed.

\spara{Range for the query budget $b$:}
We vary the number of queries,$b$, an algorithm can issue among a wide range of values.
Starting with $b<\phi(T,r)-m_0$, we gradually increase it until we see that
the performance of our algorithms stabilize (i.e., further queries do not decrease 
the reconstruction error).  Clearly, the smaller the value of $b$ the larger the 
reconstruction error of the algorithms.

\spara{Reconstruction error:} Given a ground-truth matrix $T$
and input $T_{\Omega_0}$, we evaluate the performance of a reconstruction algorithm
$\calA$, by computing the relative error of
$\wT_{\calA,{\Omega_0}}$ with respect to $T$, using the  
$\error$ function defined in Equation~\eqref{eq:error}.
This measure takes into consideration \emph{all} entries of $T$, both the observed and the unobserved.
The closer $\wT_{\calA,\Omega}$ is to $T$ the smaller the value of $\error(\wT_{\calA,\Omega})$. In general,
$\error(\wT_{\calA,\Omega})\in[0,\infty)$ and 
at perfect reconstruction $\error(\wT_{\calA,\Omega})=0$.

Although our baseline algorithms always produce a full estimate (i.e., they estimate all missing entries), {\targetedicmc} may produce only partial completions
(see Section~\ref{sec:everything} for a discussion in this).
In these cases, we assign value $0$ to the entries it does not estimate.


\begin{figure*}
\begin{center}

\subfigure[ {\movielens} ]{
\includegraphics[scale=0.33]{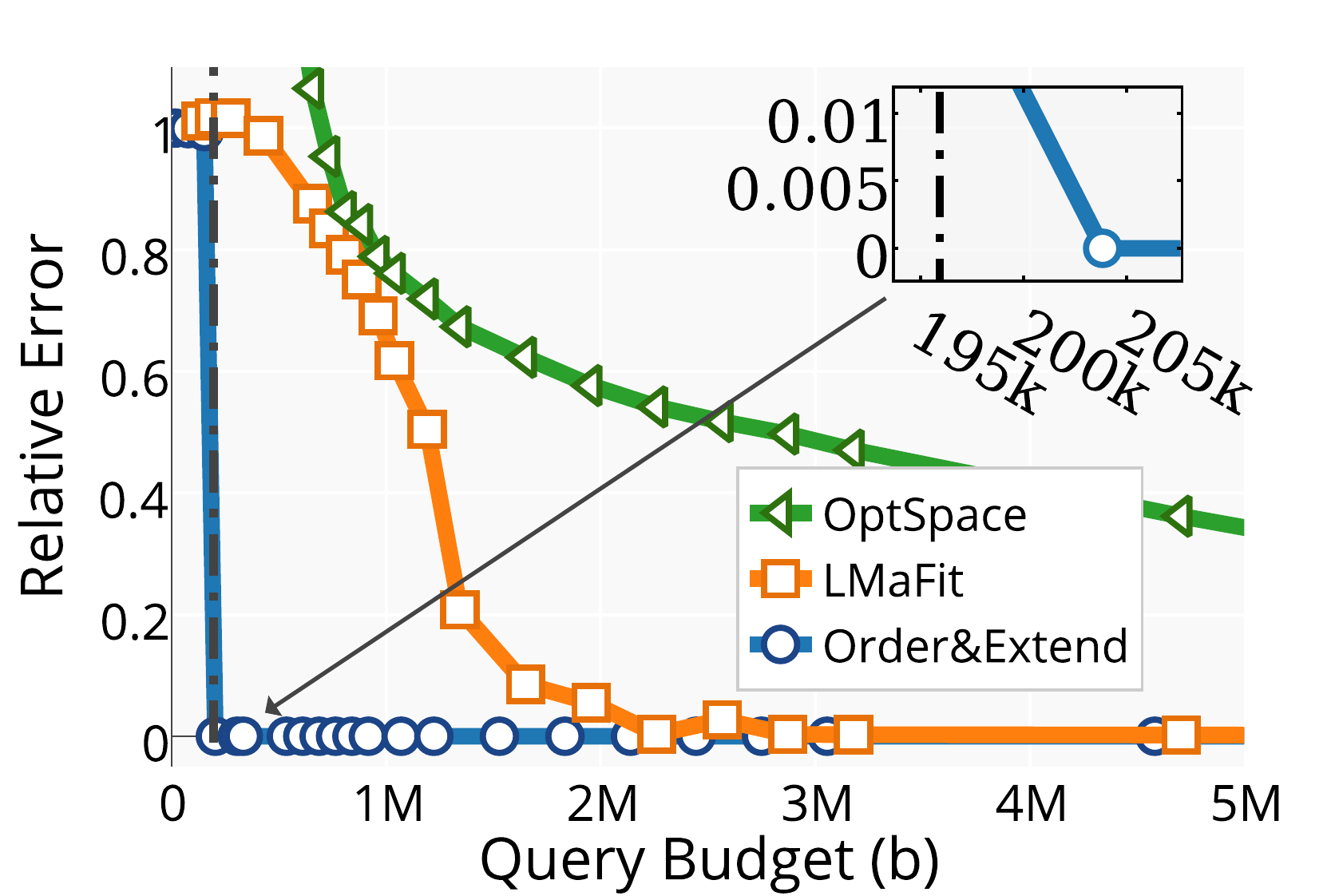}
\label{fig:movie-error}
}
\subfigure[ {\netflix} ]{
\includegraphics[scale=0.33]{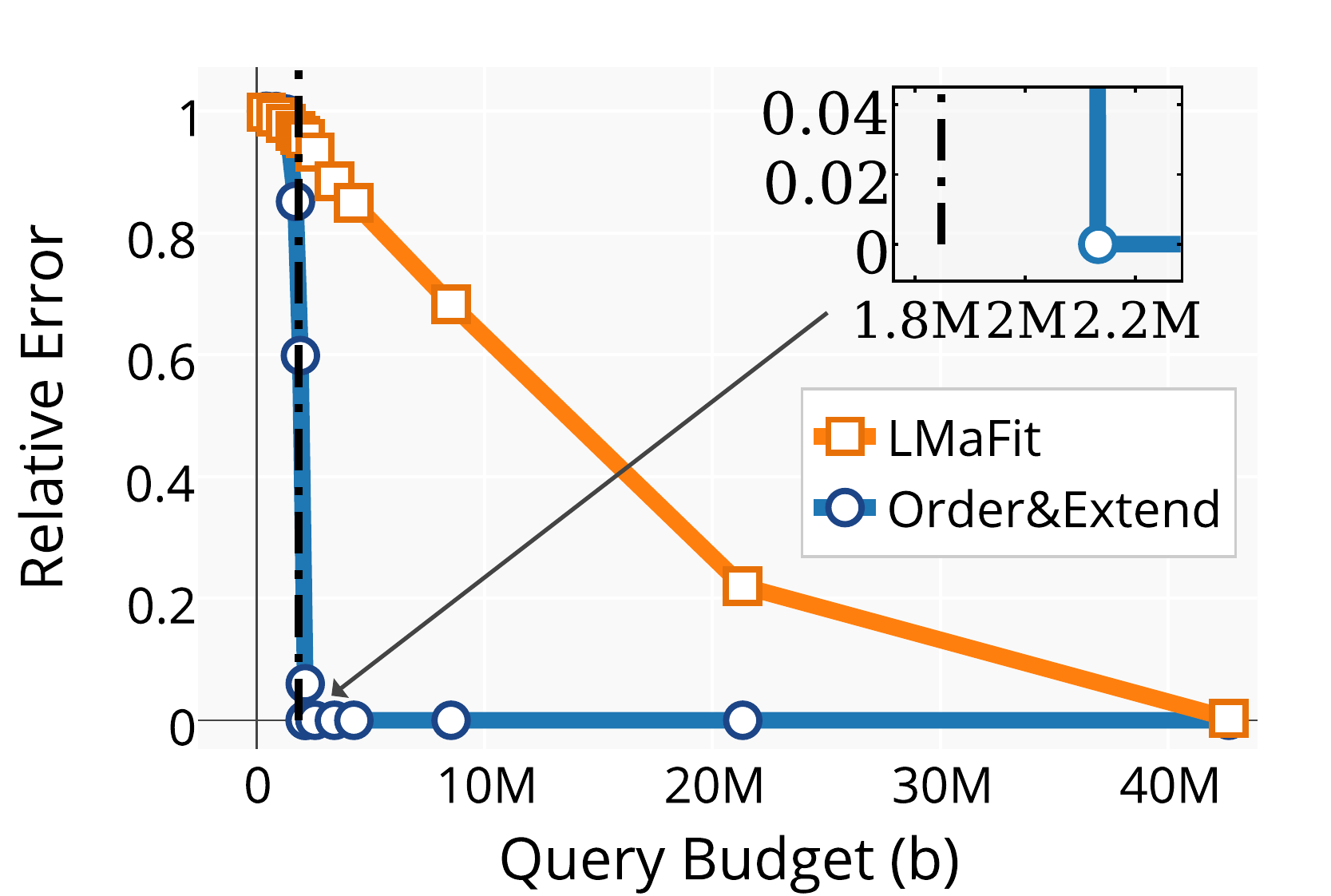}
\label{fig:netflix-error}
}
\subfigure[ {\dataset{TrafficSparse}}]{
\includegraphics[scale=0.33]{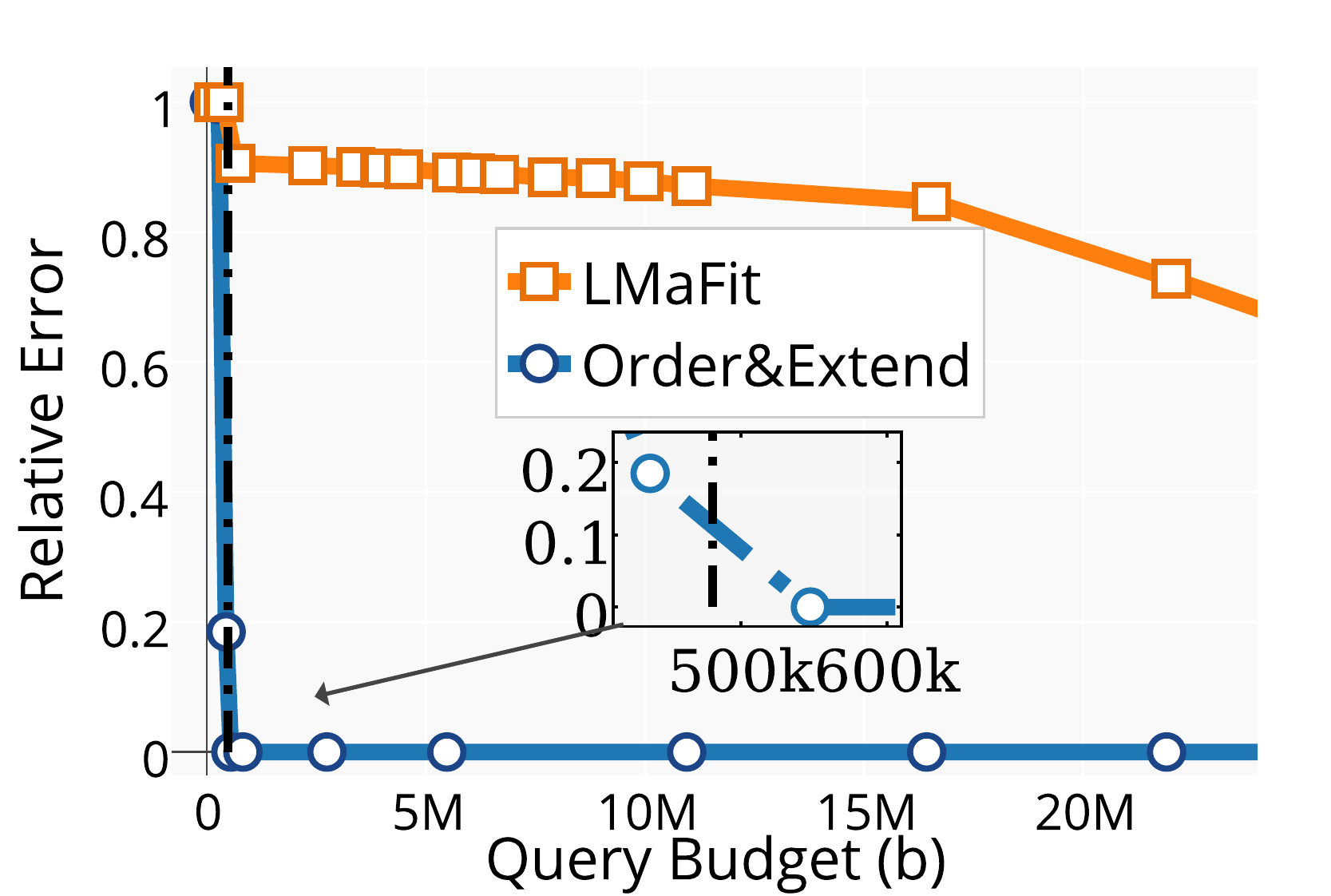}
\label{fig:traffic-error}
}
\end{center}
\caption{\label{fig:error} {\error} of completion achieved by {\targetedicmc}, {\lmafit} and {\optspace} on datasets with \emph{exact} rank; $x$-axis: query budget $b$; $y$-axis: {\error} of the completion.}
\end{figure*}

\subsection{Evaluating {\large{\targetedicmc}}}
\spara{Experiments with real noisy data:}
For our first experiment, we
use datasets for which 
we know all off the entries.
This is true for six out of our nine datasets: {\dataset{Traffic1}} ,{\dataset{Traffic2}}, {\dataset{Latency1}}, {\dataset{Latency2}}, {\dataset{Jester}}, {\dataset{Boat}}.  Note that {\dataset{Jester}} is missing $30\%$ of the entries, but we treat them as true zero-values ratings; the remaining datasets are fully known and able to be queried as needed.  As these are real datasets 
they are not exactly low rank, but plotting their singular values reveals 
that they have low effective rank. By inspecting their singular
values, we chose: $r=7$ for the {\dataset{Traffic}} and {\dataset{Latency}} datasets, $r=10$ for {\dataset{Jester}} and  $r=40$ for {\dataset{Boat}}.  

Figure~\ref{fig:errorReal} shows the results for each dataset.   The $x$-axis is the 
query budget $b$; note that while  {\lmafit} and {\optspace} always exhaust this budget, for {\targetedicmc} it is only an upper bound on the number of queries made. The $y$-axis is the ${\error}(T,\wT_{\calA,\Omega}) $. 
The black vertical line marks the number of queries needed to reach the 
critical mask size; i.e., it corresponds to 
budget  of $(\phi(T,r)-m_0)$. 
One should interpret this line as a very conservative 
lower bound on the number of queries that an optimal algorithm would need
to achieve errorless reconstruction in the absence of noise.

From the figure, we observe that 
\targetedicmc\ exhibits the lowest reconstruction error across all datasets.
Moreover, it does so with a very small number of queries, 
compared to  \lmafit\ and {\optspace}; the latter algorithms 
achieve errors of approximately the same magnitude in all datasets.
On some datasets \lmafit\ and \optspace\ come close to the relative error of \targetedicmc\ though with significantly more queries.  
For example for the {\dataset{Latency1}} dataset, {\targetedicmc} achieves error of $0.24$
with $b=2K$ queries; {\lmafit} needs $b=4K$ to exhibit an error of $0.33$, which is still
more than that of {\targetedicmc}.
In most datasets, the differences are even more pronounced; 
e.g., for {\dataset{Traffic2}}, {\targetedicmc} achieves a relative 
error of $0.50$ with about 
$b=13K$ queries; {\optspace} 
and {\lmafit} achieve error of more than $0.8$ even after $b=26K$ queries.
Such large differences between {\targetedicmc} and the baselines appear in all 
datasets, but {\dataset{Boat}}. For that dataset, {\targetedicmc} is still better,
but not as significantly as in other cases -- likely an indication that the dataset is more noisy. 
We also point out that the value of $b$ for which the relative error
of {\targetedicmc} exhibits a significant drop is much closer to the 
indicated lower bound by the black vertical line.  Again this phenomenon is not so evident
for  {\dataset{Boat}} probably because this datasets is further away from being 
low rank. 


\spara{Extremely sparse real-world data:}
For the purpose of experimentation our algorithm needs to have
access to all the entries of 
the ground truth matrix $T$ -- in order to be able to 
reveal the values of the queried entries.
 Unfortunately,  the {\dataset{Movielens}},  {\dataset{Netflix}}, and {\dataset{TrafficSparse}} datasets consist mostly of missing entries, therefore we cannot query the majority of them.
To be able to experiment with these datasets, we overcome this issue 
by approximating each dataset with its closest rank $r$ matrix $T_r$. The approximation is obtained by first assigning $0$ to all missing entries of the observed $T$, and then taking the singular value decomposition and setting all but the largest $r$ singular values to zero. This trick grants us the ability to study the special case discussed in Section~\ref{sec:problem} where the matrix is of exact rank $r$.

\newpage
Using $r=40$, the results for these datasets are depicted in Figure~\ref{fig:error} with the same axes and vertical line as in Figure~\ref{fig:errorReal}.  Again, we observe a clear dominance of {\targetedicmc}.  In this case the differences in the relative error
it achieves are much more striking. Moreover,
{\targetedicmc} achieves almost $0$ relative error for extremely small
number of queries $b$; in fact the error of {\targetedicmc}
consistently drops to an extremely small value for $b$ very close to the 
lower bound of the optimal algorithm (as marked by the black vertical line shown in the plot).
On the other hand {\lmafit} and {\optspace} are far from exhibiting such a behavior.  This signals that 
{\targetedicmc} devises a querying strategy that is almost optimal.
Interestingly the performance of \optspace\ changes dramatically in these cases as compared to the approximate rank datasets. In fact on {\dataset{TrafficSparse}} and {\dataset{Netflix}} the error is so high it does not appear on the plot.

Note that the striking superiority of {\targetedicmc} in the case of exact-rank
matrices is consistent across all datasets we considered, including others not shown 
here.

\begin{figure}[H]
\begin{center}
\includegraphics[width=8cm,height=4cm]{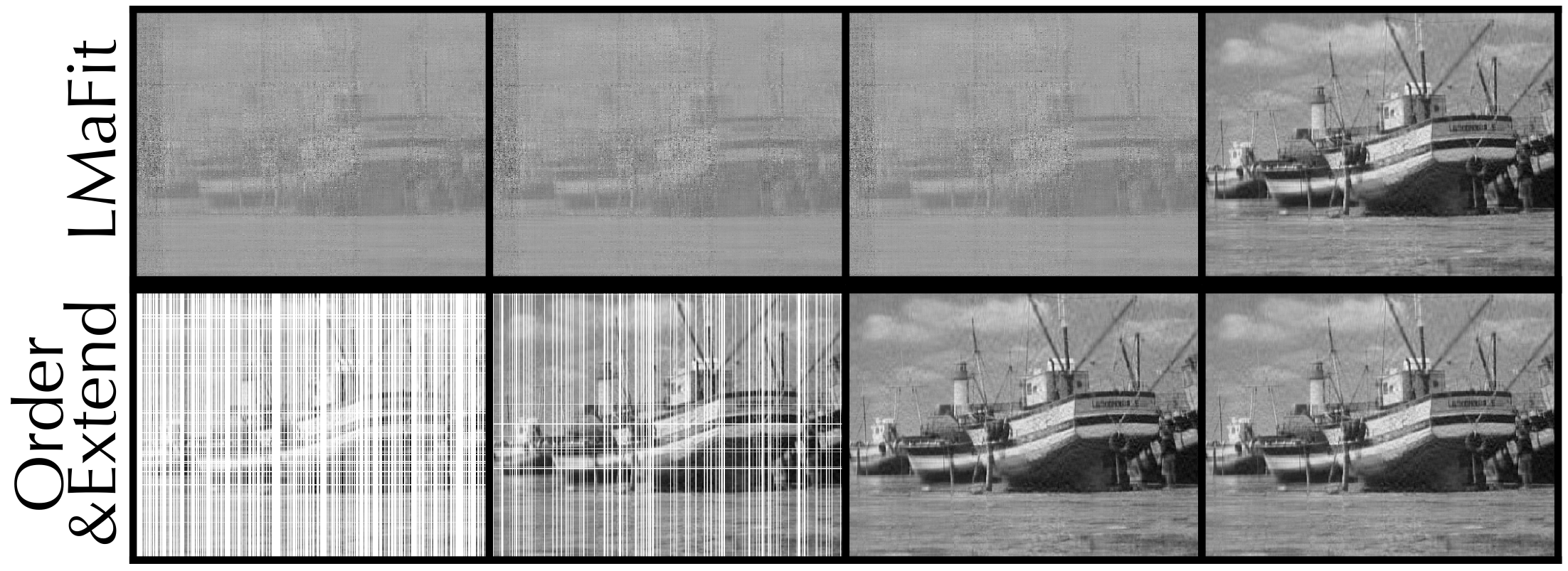} 
\end{center}\vspace{-.3cm}
\caption{\label{fig:visual} Recovery process using {\lmafit}, and {\targetedicmc}.  Each column is a particular $b$, increasing from left to right. }
\end{figure}


\vspace{-.2cm}
\spara{Running times:}
Though the algorithmic composition is quite different, we 
give some indicative running times for 
our algorithm as well as {\lmafit} and {\optspace}.
For example, in the 
{\dataset{Netflix}} dataset the running times were in the order of
 $11\, 000$ seconds for {\lmafit}, $80\, 000$ seconds for {\targetedicmc}, and $200\, 000$ seconds for {\optspace}.  These numbers  indicate that {\targetedicmc}
 is efficient despite the fact that in addition to matrix completion it also 
 identifies the right queries; the running times of {\lmafit} and {\optspace}
 simply correspond to running a single completion on the extended mask 
 that is randomly formed. 
Note that these running times are computed using an unoptimized and serial 
implementation of our algorithm; improvements can be achieved easily e.g., by parallelizing
the local condition number computations.

\spara{Partial completion of {\targetedicmc}:} 
As a final experiment, we provide anecdotal evidence that demonstrates the 
difference in the philosophy behind {\targetedicmc} and other completion algorithms.
  Figure~\ref{fig:visual} provides a visual comparison of the recovery process of \targetedicmc\ and {\lmafit} for different values of query budget $b$.
For small values of $b$, 
\targetedicmc\ does not 
have the sufficient information to resolve all incomplete and unstable systems.  Therefore the algorithm does not estimate the entries of $T$ corresponding to these systems, which renders the white areas in the two left-most images of \targetedicmc\ . In contrast {\lmafit} outputs full estimates, though with significant error.  This can be seen by incremental sharpening of the image, compared to the piece-by-piece reconstruction of {\targetedicmc}.


\subsection{Discussion}
\label{sec:discussion}
Here we discuss some alternatives we have experimented with, but omitted due to significantly poorer performance.

\spara{Alternative querying strategies:} 
\targetedicmc\ uses a rather intricate strategy for choosing 
its queries to $T$.  A natural question is whether a simpler strategy would 
be sufficient.  
To address this we experimented with versions of \seqS\ that  
considered the same order as {\targetedicmc}
but when stuck with a problematic system they queried either randomly, or with probability proportional (or inversely proportional) to the number of observed 
entries in a cell's row or column.
All these variants were significantly and consistently worse than 
the results we reported above.

\spara{Condition number:} Instead of detecting unstable systems using the local 
condition number we also experimented
with a modified version of 
{\targetedicmc}, which characterized 
a system $A_xy=t$ as unstable if its condition number $\kappa(A_x)$ was above a threshold.
For values of threshold between $5$ and $100$ the results were 
consistently and significantly worse than the results of {\targetedicmc} that we report here,
both in terms of queries and in terms of error.
Further, there was no threshold of the condition number that would perform comparably to {\targetedicmc} for any dataset.

\


\vspace{-.7cm}
\section{Conclusions}\label{sec:conclusions}
In this paper we posed the {\minrev} problem, an active version of
matrix completion,
and designed an efficient algorithm for solving it.
Our algorithm, which we call {\targetedicmc},
approaches this problem by viewing querying and completion
as two interrelated tasks and optimizing for both simultaneously.
In designing {\targetedicmc} we relied on 
a view of matrix 
completion as the solution of a sequence of linear systems, 
in which the solutions of earlier systems become the inputs 
for later systems.  In this process, reconstruction error depends both on the order
in which systems are solved and on the stability of each solved system.
Therefore, a key idea of {\targetedicmc} is to find an ordering for the systems
in which as many as possible give good estimates of the unobserved entries. 
However, even in the perfect order problematic systems arise; {\targetedicmc}
employs a set of techniques for detecting these systems and alleviating
them by querying a small number of additional entries from the true matrix.
In a wide set of experiments with real data we demonstrated the efficiency of our algorithm and 
its superiority both in terms of the number of queries it makes, and the
error of the reconstructed matrices it outputs.

\spara{Acknowledgments:} This research was supported in part by NSF grants CNS-1018266, 
CNS-1012910, 
IIS-1421759, IIS-1218437,
CAREER-1253393,
IIS-1320542, and
IIP-1430145.  We also thank the anonymous reviewers for their valuable
comments and suggestions.

%

%


\bibliographystyle{abbrv}
\bibliography{icmc}

\end{document}